\documentclass{bmvc2k}

\usepackage{xspace}
\usepackage{algorithm}
\usepackage{algpseudocode}
\usepackage{amsmath,amssymb}
\usepackage{multirow}
\usepackage{graphicx}
\usepackage{multicol}
\usepackage{makecell}
\usepackage{comment}

\title{Unsupervised Video Continual Learning via Non-Parametric Deep Embedded Clustering}

\addauthor{Nattapong Kurpukdee}{nattapong.kurpukdee@york.ac.uk}{1}
\addauthor{Adrian G. Bors}{adrian.bors@york.ac.uk}{2}

\addinstitution{
Department of Computer Science, \\
 University of York,\\
 York, YO10 5GH, UK
}

\runninghead{Kurpukdee and Bors}{Unsupervised Video Continual Learning}

\begin{document}

\maketitle

\begin{abstract}
We propose a realistic scenario for the unsupervised video learning where neither task boundaries nor labels are provided when learning a succession of tasks. We also provide a non-parametric learning solution for the under-explored problem of unsupervised video continual learning. Videos represent a complex and rich spatio-temporal media information, widely used in many applications, but which have not been sufficiently explored in unsupervised continual learning. Prior studies have only focused on supervised continual learning, relying on the knowledge of labels and task boundaries, while having labeled data is costly and not practical. To address this gap, we study the unsupervised video continual learning (uVCL). uVCL raises more challenges due to the additional computational and memory requirements of processing videos when compared to images.
We introduce a general benchmark experimental protocol for uVCL by considering the learning of unstructured video data categories during each task. We propose to use the Kernel Density Estimation (KDE) of deep embedded video features extracted by unsupervised video transformer networks as a non-parametric probabilistic representation of the data. We introduce a novelty detection criterion for the incoming new task data, dynamically enabling the expansion of memory clusters, aiming to capture new knowledge when learning a succession of tasks. We leverage the use of transfer learning from the previous tasks as an initial state for the knowledge transfer to the current learning task. We found that the proposed methodology substantially enhances the performance of the model when successively learning many tasks. We perform in-depth evaluations on three standard video action recognition datasets, including UCF101, HMDB51, and Something-to-Something V2, without using any labels or class boundaries. 
\end{abstract}

\section{Introduction}
\label{sec:intro}

Unsupervised Continual Learning (UCL) aims to progressively learn from unlabeled data by finding associations based on certain criteria while addressing catastrophic forgetting.
Usually, groupings of data are made according to their statistical similarity.  A key challenge to this process is that of being able to preserve what was learned in the past, representing the stability, while also having the ability to learn novel information, corresponding to plasticity. The trade-off between stability and plasticity in unsupervised video learning represents a challenging endeavor.

Most unsupervised class-incremental learning approaches developed for the image domain \cite{lin2022prototype, liang2020we, wei2024class, da2022unsupervised, hao2023dual, reddy2024unsupervised} focus on aligning unlabeled data with those from categories derived from labeled source data. However, these methods rely on large supervised models and make unrealistic assumptions such that all given unlabeled data represent novel information, without considering overlaps with previously learned data. Furthermore, they require predefined cluster boundaries, which limit their applicability to real-world scenarios.

Unlike image-based UCL models, the video domain received very limited attention in continual learning studies. In real-world situations, incoming unlabeled data often consists of data sourced from different probabilistic representations, corresponding to mixed sets of data categories overlapping with each other as well as with the previously learned information. In such cases, a challenge is represented by the insufficient amount of data available to fully train the model and by the differences in the amount of such data from different categories. In the method proposed here we initially extract feature sets using a video transformer \cite{wang2023videomae}. Then, we successively organize the extracted sets of features into clusters, representing the statistical distribution characterizing the learned data. During the learning of each task, the model continuously associates new feature sets with existing clusters while also creating new clusters according to a novelty criterion, optimizing both memory and time efficiency. We consider a non-parametric clustering method by adapting the mean-shift algorithm \cite{MeanShift,1000236,KDE_BorNas09} as a continual representation through the Kernel Density Estimation (KDE) representation of video data. 

In this paper, we address the real-world challenge of unsupervised continual learning for video, where neither task boundaries nor class boundaries are provided to the learner. We propose a simple yet effective and practical approach entitled the unsupervised Video Continual Learning based on Kernel Density Estimation (uVCL-KDE). 

Our main contributions are as follows~:
\begin{itemize}
 \item We explore a non-parametric continual learning setting through the proposed uVCL-KDE, by grouping data based on their kernel-density representation affinities.
  \item We propose to use the mean-shift method, for defining sets of clusters when applying uVCL-KDE on video features in a continual learning setting. We also extend to the uVCL-KDE-RBF by adding a linear mapping on top of the clustering, as in the Radial Basis Function (RBF) networks.
  \item We introduce a benchmark evaluation protocol to facilitate a realistic assessment framework for future research and provide an extensive experimental analysis of the effectiveness of each component in our proposed approach.
\end{itemize}

\section{Related Work}
\label{sec:relatedwork}

Continual or lifelong learning is characteristic to all living beings allowing them to adapt in various life situations. However, AI systems suffer from catastrophic forgetting when they are retrained on new datasets and have a very low probability of fulfilling the tasks learned in the past.
In this section, we begin by reviewing existing research on supervised continual learning, highlighting key challenges. Following this, we examine various unsupervised continual learning and their potential for real-world applications.

\subsection{Supervised Continual Learning}

In Supervised Continual Learning (SCL), models are trained sequentially on a series of $k$ tasks $\{ \tau_0, \tau_1, \ldots, \tau_k \}$, where each task involves learning a set of data and its corresponding labels.  SCL approaches are generally categorized into those based on regularization, architecture expansion and memory-based methods. Regularization-based approaches use some specific terms in the loss function in order to reduce catastrophic forgetting \citep{EWC,aljundi2018memory}. Meanwhile, expansion architecture models add new neurons, layers or entire modules in order to enable the learning of new tasks \cite{CompactingPicking}. Memory-based methods mitigate catastrophic forgetting by retaining a limited subset of training data from previously learned tasks $\{ \tau_0, \tau_1, \ldots, \tau_{k-1} \}$ in a memory buffer. Then they draw samples from the memory buffer when learning a new task $\tau_k$.

Most existing video supervised continual learning (VSCL) models are adaptations of methods initially developed for image continual learning  \citep{iCaRL,aljundi2018memory, BIC,cermelli2020modeling,2020podnet, FrameMaker,CIL-StrongMod, L2P,LLT-S,CIL-Survey,pei2023space,LGAA, kong2023trust}. Some image-based VSCL approaches, such as the Incremental Classifier and Representation Learning (iCaRL) \citep{iCaRL}, and Bias Correction (BiC) \citep{BIC}, have been directly extended to the VSCL models, \citep{CIL-FOR-ACTION-CLASSIFICATION,CIL-FOR-ACTION-RECOGNITION-IN-VIDEOS,vCLIMB}. Most  models use memory buffers to store videos from previously learned classes during continual learning, together with their labels, aiming to address catastrophic forgetting \cite{NatICIP2024}.
Other methods, specifically proposed for video SCL \citep{CIL-FOR-ACTION-CLASSIFICATION,CIL-FOR-ACTION-RECOGNITION-IN-VIDEOS, maraghi2022class, vCLIMB, pei2023space}, focus on mitigating forgetting in video-based tasks by using memory buffers or prompts to retain the knowledge of previously learnt classes. Many video SCL models rely on Convolution Neural Networks (CNNs) as their backbones. Recently, a promising direction of research is represented by the integration of large language models (LLMs) and vision models for video SCL \cite{pei2023space,PIVOT_villa}. However, these methods are still limited in their real-world applicability due to their reliance on costly human annotation and labeling.

\subsection{Unsupervised Continual Learning}

Unsupervised learning in the image domain is a rapidly growing research area, with various methods \cite{caron2018deep,wu2018unsupervised} leveraging visual features for learning without any labeled information. Similarly, the unsupervised learning in videos is increasingly gaining attention, with Zhuang {\em et al.} \cite{zhuang2020unsupervised} introducing a two-pathway approach for unsupervised video learning. Meanwhile, unsupervised Continual Learning (UCL) in the image domain has  been explored in several studies, such as \cite{rao2019continual, cheng2023cucl, shin2017continual, he2021unsupervised, taufique2022unsupervised, madaan2021representational, wu2018unsupervised, lin2022prototype}. In these models, pseudo-labels are used to replace human annotations for learning new, non-overlapping categories of image data. To address the problem of catastrophic forgetting, some methods use the Deep Generative Replay (DGR) replay training. Additionally, simple classifiers like K-Nearest Neighbors (KNN) are employed in the latent space for unsupervised data assignment. 

The field of unsupervised continual learning in video domain remains under-explored. While unsupervised domain adaptation has been studied for both images and videos in various applications \cite{liang2020we,wei2024class, da2022unsupervised,hao2023dual,reddy2024unsupervised}, these models typically depend on a pre-trained, supervised source model, while adapting the unsupervised target data to the identified primary source representations. 

Unlike the previous studies that consider either class or category incremental learning settings, in this paper we propose a simple yet effective framework, consisting of learning from the data of multiple mixed categories in an unsupervised way. The proposed approach relies on the kernel density-based data representation in the video feature space. The resulting peaks in the non-parametric representation provided by the Kernel Density Estimation (KDE) of the feature space are used as data representation attributes to store and then replay past information during continual learning.

\section{Problem Setup}
\label{sec:problem_setup}

In this paper, we study unsupervised video continual learning, aiming to learn and structure a data space ${\cal H}$, given a sequence of $K$ tasks, $\{\tau_1, \tau_2, \dots, \tau_K\}$. During each task $\tau_k$, $k=1,\ldots,K$ a set of video data is provided, denoted as 
$ \tau_k = \{ {\bf v}_i \}_{i=1}^{n_k}$, where ${\bf v}_i$ represents $i$-th video sample and $n_k$ is the number of videos to be learnt during task $\tau_k$, with a total number of training data as $n=\sum_{k=1}^K n_k$. In this study we consider the challenging situation of learning unsupervised tasks, where there are no labels for the training set. We also assume no pre-defined structure or categorization in the video data. Our objective is to define a labeling function $f(\cdot)$, parameterized by a deep learning network, that assigns pseudo-labels to the data, $\tilde{y}_i^j =  f({\bf v}_i)$, where we assume that the ideal label, unknown to $f(\cdot)$ is $y_i^j$, where $j$ is the label's identifier. The pseudo-labels assignment is performed according to the video's feature properties and the characteristics of the labeling function $f(\cdot)$. 

Under the most general, yet realistic, setting, each task $\tau_k$ consists of mixed class/category data, corresponding to a mixture of distributions, where $\tau_k = \{{\bf v}_i, y_j | i=1,\ldots,n_k, j=1,\cdots,$ $m_k\}$, where  $y_j$ represents true labels, unknown in the unsupervised learning system $f(\cdot)$ and $m_k$ is the number of data categories provided at $\tau_k$. 
The data learnt at task $\tau_k$ may or may not overlap statistically with previously learned data, without actually providing explicit class boundaries for any learnt datum. This presents a more realistic, yet challenging problem, that was not tackled before, as the model must learn at each  step from completely unstructured data while aiming to form semantically meaningful data associations defined by $f(\cdot)$. Our aim is to create a series of representative clusters, with each cluster characterized by a pseudo-label $\tilde{y}_i^j$, $j=1,\ldots,l_k$, where $l_k$ represents the number of identified clusters at task $\tau_k$. 

In the proposed unsupervised continual learning, in order  to mitigate catastrophic forgetting, we consider storing a small set of video features
${\cal P}_k= \bigcup_{j=1}^{l_k}  \{{\bf v}_i, \tilde{y}_j\}_{i=1}^{n_k}$, as exemplars associated with each cluster $j$, defined at task $\tau_k$. 
These data are then reused during each subsequent task learning, while more clusters are added. Unlike supervised class-incremental learning approaches \cite{iCaRL,aljundi2018memory,BIC,cermelli2020modeling, 2020podnet, FrameMaker,CIL-StrongMod, L2P,LLT-S,CIL-Survey,pei2023space,LGAA, kong2023trust,vCLIMB}, which operate with fixed class increments, $l_1=l_2=\ldots=l_K$, our approach does not require to predefine the number of classes for each incremental step. Instead, the number of classes dynamically increases throughout the continual learning process, reflecting the non-stationary nature of the real-world data.

\section{Method}
\label{sec:method}

\begin{figure}
  \centering
  \includegraphics[width=\linewidth]{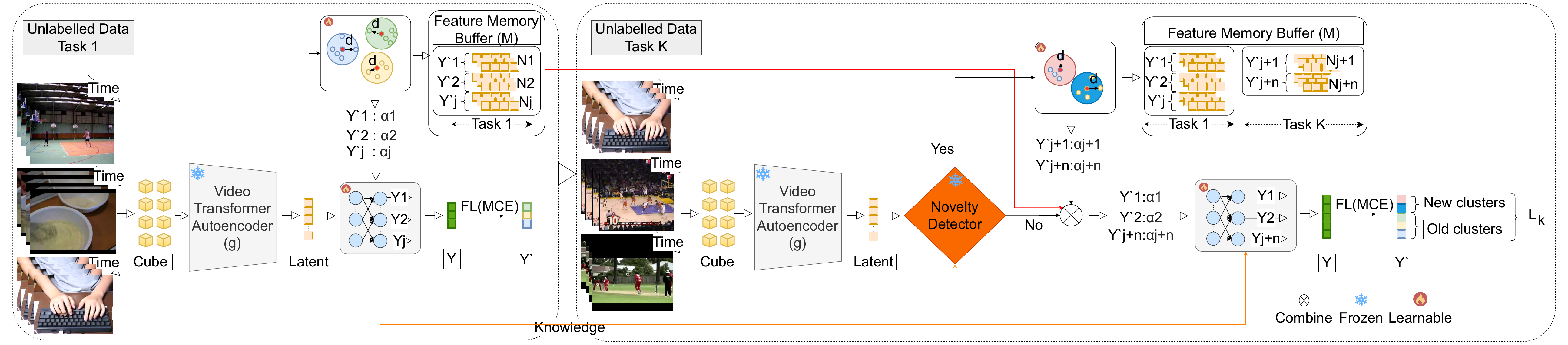}
  \caption{Overview of the proposed unsupervised video continual learning based on the Kernel Density Estimation (uVCL-KDE).} 
  \label{fig:proposed}
\end{figure}

We propose a simple yet effective method for unsupervised video continual learning, based on non-parametric deep-embedded cluster assignments. The overall procedure for unsupervised continual learning of a sequence of video tasks $\{ \tau_1, \tau_2, \dots, \tau_k \}$ is outlined in Fig.~\ref{fig:proposed}. In the following, we describe the proposed approach, which includes feature extraction using a video transformer, non-parametric deep cluster embedding along the continual learning process and the memorization of features for the memory replay during future task learning.

\subsection{Feature extraction}
\label{Features}

Each task involves using a video auto-encoder transformer network for extracting video features, $\{{\bf x}_{k,i} = g ({\bf v}_i)\}_{i=1}^{n_k}$, using a pre-trained auto-encoder video transformer network, considered of size $|{\bf x}_{k,i}|$, for any task $\{ \tau_k | k=1,\ldots,K\}$ and $i=1,\ldots,n_k$. The network $g( \cdot)$ is used to extract the features when learning all tasks $\{ \tau_k | k=1,\ldots,K\}$, without being retrained, thus ensuring a consistent feature space over the entire data space.
These features are then grouped into a number of clusters through a deep clustering algorithm, described in the following.

\subsection{KDE-based Deep Embedded Clustering}
\label{sec:deep_embedded_clustering}

In this paper we propose the unsupervised Video Continual Learning using Kernel Density Estimation (KDE), namely uVCL-KDE. The proposed methodology relies on the online non-parametric deep embedded clustering strategy. 
The video feature data are organized by the Mean-shift \cite{MeanShift,1000236,KDE_BorNas09}, which is a dynamic data-representation KDE-based clustering method which does not require to know the number of clusters. The data representation in the KDE is given by considering a kernel function centered on each sample and calculating the resulting probability density function (pdf)~:
\begin{equation}
\begin{aligned}
   \hat{f}({\bf x}) = \sum_{i=1}^{n_k} {\cal K}({\bf x}-{\bf x}_i) = \sum_{i=1}^{n_k} {\cal K} \left( \frac{\left\|{\bf x} - {\bf x}_i \right\|^2}{h^2} \right)
  \end{aligned}
 \label{eq:meanshift}
\end{equation}
where ${\bf x}_i`$, $i=1,\ldots,n$ represents the feature vector of the input data, where we consider ${\cal K}$ depending on a bandwith $h$, with each kernel centered at a data sample. $h$ can influence the number of peaks in the resulting probability density function (pdf) representation \cite{KDE_BorNas09}. 

When considering the Gaussian kernel, with its cluster center $\mbox{\boldmath$\mu$}_j$ as the kernel center while $h$ corresponds to the standard deviation, the Mean-shift, adaptively moves its cluster centers towards the peaks in the pdf representation, like the one from Eq.~\eqref{eq:meanshift}. Then, when considering ${\cal K}$ as a Gaussian kernel \cite{1000236} in Eq.~\eqref{eq:meanshift} and  differentiating this pdf representation, we can iteratively calculate the mean-shift as~:
\begin{equation}
M (\mbox{\boldmath$\mu$}_j^t) = \frac{\sum_{i=1}^{n_k} \left( - {\bf x}_i  
\frac{ \|\mbox{\boldmath$\mu$}_j^t - {\bf x}_i \|^2}{2h^2} \right) }{ \sum_{i=1}^{n_k} \left( -  \frac{ \|\mbox{\boldmath$\mu$}_j^t - {\bf x}_i \|^2}{2h^2} \right)}  - \mbox{\boldmath$\mu$}_j^{t-1},
\label{MeanShift}
\end{equation}
where $\mbox{\boldmath$\mu$}_j^t$ is the cluster center found at the $t$-th iteration of the Mean-shift algorithm. The Mean-shift is iteratively used to update the mean as $\mbox{\boldmath$\mu$}_j^{t+1}$, and then is recalculated until the cluster centers are found $\mbox{\boldmath$\mu$}_j = \mbox{\boldmath$\mu$}_j^t$ when $\mbox{\boldmath$\mu$}_j^t \approx \mbox{\boldmath$\mu$}_j^{t-1} $. After finding the peaks from the KDE representation corresponding to all the data from the given task $\{ \tau_k| i=1,\ldots,n_k \} $ two cluster candidates are considered as distinct if there is a local minima on the line that joins them, while otherwise the two clusters are merged. 

Eventually, a cluster is associated with each peak in the resulting KDE and the peaks are found through the mean-shift, as described above. Consequently, data are associated with the peaks and clusters. In order to avoid forgetting in the unsupervised video continual learning, after learning each task, a certain number of video features, are stored in memory buffers in order to be used for future training. We assign a memory buffer ${\cal M}_j$, $j=1,\ldots,l_k$ for each peak of the KDE, considered as defining a cluster in the KDE representation. When proceeding to the next task $\tau_{k+1}$, all the data from the memory are combined and used together with the new data provided with the task, forming an updated KDE landscape. 
After iterating through equations \eqref{eq:meanshift} and \eqref{MeanShift} new clusters are formed when novel data are identified, resulting in a probabilistic representation that adapts to the novel data, while also preserving the knowledge accumulated during the learning of all tasks $\{ \tau_1, \tau_2, \dots, \tau_k, \tau_{k+1} \}$.

\subsection{Linear cluster self-allocation}
\label{RBF-linear}

All video data  ${\bf x}_{k,i}$ associated uniquely to each cluster, are assigned with a pseudo-label $\Tilde{y}_{k,i}$.  Such data allocations, defined by the centers $\mbox{\boldmath$\mu$}_{k,j}$, can be seamlessly integrated with regularization-based methods, such as knowledge distillation loss or other approaches. In this context, a multi-class cross-entropy loss is applied to learn the cluster assignments for the data ${\bf x}_{k,i}$, with such pseudo-labels being akin to labels typically used in supervised settings. Here, $\Tilde{y}_j$, $j=1,\ldots,L_K$ represent the $L_K$-cluster assignments and these are used as targets within a linear classifier, like in a Radial Basis Functions (RBF) network  \cite{RBF}. The cluster assignments  $\Tilde{y}_j$ are used as training labels in a multi-class classification task. The classifier then outputs class probabilities using softmax normalization, as in the following~: 
\begin{equation}
  \sigma(\Tilde{y}_{i,j} ) = \frac{\exp(\Tilde{y}_{i,j})}{\sum_{j=1}^{L_K} \exp(\Tilde{y}_{i,j})} \ \ \ \mbox{for} \ i=1,2,\dots,n,
  \label{eq:softmax}
\end{equation}
where, $\Tilde{y}_{i,j}$ is the vector of raw outputs from the neural network, and $\sigma(\Tilde{y}_{i,j} )$ is the softmax output corresponding to the probability that the input belongs to class $i \in L_K$ and $L_K$ is the number of pseudo-classes, given by the number of clusters. 

This method, which trains a linear layer on top of the clusters inferred from the KDE representation, akin to the Radial Basis Function (RBF) Networks \cite{RBF}, is named uVCL-KDE-RBF.
For training the last layer of uVCL-KDE-RBF, we use the multi-class cross-entropy (MCE) loss~:
\begin{equation}
  MCE = -\sum_{j=1}^{L_K} \overline{y}_{o,j}  \log(p_{o,j}),
\label{eq:MCE}
\end{equation}
where $L_K$ represents the classes defined by the pseudo-clusters, $\overline{y}$ is a binary indicator of 0 or 1 indicating whether the class label $j$ is the correct classification label for observation $o$. $p$ is the predicted probability observation $o$ for the class $j$. 

A challenging aspect in the unsupervised continuous learning is the presence of unbalance in the amount of different data categories. Consequently, we address such imbalances during training by employing the Focal Loss for weighting the contribution of each cluster \cite{FOCALLOSS}. We then modify the MCE using Focal Loss (FL) from \cite{FOCALLOSS}, by replacing the classes with the pseudo-labeled clusters, as~:
\begin{equation}
  FL(MCE) = \alpha_j * (1 - \exp{(-MCE)} )^\gamma *MCE,
  \label{eq:class_weight_facal_loss}
\end{equation}
where $\alpha_j$ is the pseudo-cluster balance weight, $j=1,\ldots,L_K$,  and $\gamma=2$ is a modulating factor for the multi-class cross-entropy loss. 

\subsection{Novelty detector and cluster augmentation }

New clusters are defined when the new data, provided in the tasks from the sequence being learned, indicates completely different information from that already known by the uVCL-KDE, according to~:
\begin{equation}
   \arg\min_{j=1}^{L_{K-1}}  d_j ({\bf x}_{k,i}) = \parallel g({\bf x}_{k,i} ) - \mbox{\boldmath$\mu$}_{k,j} \parallel  > \Theta_1,
   \label{eq:Theta1}
\end{equation}
where $\Theta_1$ is a threshold defining new clusters.
We estimate $\Theta_1$ using the data from the first task, as the maximum of the distances between each two existing cluster centers.

In the case of the uVCL-KDE-RBF we consider the probability $\sigma(\Tilde{y}_{i,j} )$  from Eq.~\eqref{eq:softmax} for the data learnt by all tasks $\{ \tau_j| j=1, \ldots, k-1 \}$. We consider a maximum probability $\sigma(\Tilde{y}_{i,j} )$ for defining an existing cluster. For a data sample ${\bf x}_{k,i} \in \tau_k$, we consider a new cluster, if after evaluating \eqref{eq:softmax}, we have~:
\begin{equation}
   \arg\max_{j=1}^{L_{K-1}} \sigma(\Tilde{y}_{i,j} )  < \Theta_2, 
   \label{eq:theta_2}
\end{equation}
where $\Theta_2$ defines new clusters.
The memory management strategy associated with all clusters is explained in Appendix A from the Supplementary Material (SM).

\section{Experimental Results}
\label{sec:experimeant}

In this section, we evaluate the proposed unsupervised video continual learning methodology. We consider   UCF101 \cite{UCF101}, HMDB51 \cite{HMDB51} and  Something-Something V2 (SSv2) \cite{SSV2} datasets, after dropping all class labels, in order to use the data for unsupervised learning. Details about the datasets are provided in Appendix B from SM.

\subsection{Implementation Details} 

We implement uVCL-KDE and uVCL-KDE-RBF using PyTorch \cite{paszke2019pytorch} and the Adam optimizer \cite{kingma2014adam} with a learning rate of 0.001, considering a single NVIDIA GeForce GTX 1080 Ti 11GB GPU. For each dataset, models are trained for up to 50 epochs, using a batch size of 8 videos. The model is optimized using the Focal Loss (FL) for balancing different video categories, as in Eq.~\eqref{eq:class_weight_facal_loss} considering $\gamma=2$, and using the Scikit-learn's framework \cite{scikit-learn} for the implementation. We utilize the Scaling Video Masked Autoencoders with Dual Masking (VideoMAE V2) \cite{wang2023videomae} as the auto-encoder video transformer for implementing the feature extractor $g(\cdot)$ to capture spatio-temporal features, as illustrated in Fig.~\ref{fig:proposed}. This transformer model is pre-trained on the Kinetics-700 dataset \cite{kay2017kinetics} without any label information. The input videos are composed of $16$ frames, of size $224 \times 224 \times 3$ pixels. During pre-processing, the video frames are center-cropped and their pixels re-scaled to the range [0.0, 1.0], then normalized using a mean of  $[0.485, 0.456, 0.406]$ and a standard deviation of  $[0.229, 0.224, 0.225]$. The extracted output features for each video $i$ at task $k$ has $|{\bf x}_{k,i}| = 1024$ channels. At the end of each task learning, we store the features for $N=20$ videos per each cluster in the memory buffers ${\cal M}_i$. The baselines used in the experiments are described in Appendix C from SM.

{\bf Evaluation Metrics.} We adapt the protocol used in the unsupervised settings from \cite{caron2018deep,van2020scan}, evaluating the cluster accuracy (CAcc), used for the unsupervised continual learning for images \cite{he2021unsupervised}. Then, we evaluate the average unsupervised continual learning accuracy over all the training tasks, including the final task accuracy (ACAcc) as in \cite{lopez2017gradient, vCLIMB}. More details about the evaluation metrics used, including the Forward Forgetting (FWF) and the Backward Forgetting (BWF) are provided in Appendix D from SM.
A large positive Forward Forgetting is also known as catastrophic forgetting.

\begin{figure}
\begin{center}
\begin{tabular}{ccc}
\includegraphics[width=0.3\textwidth]{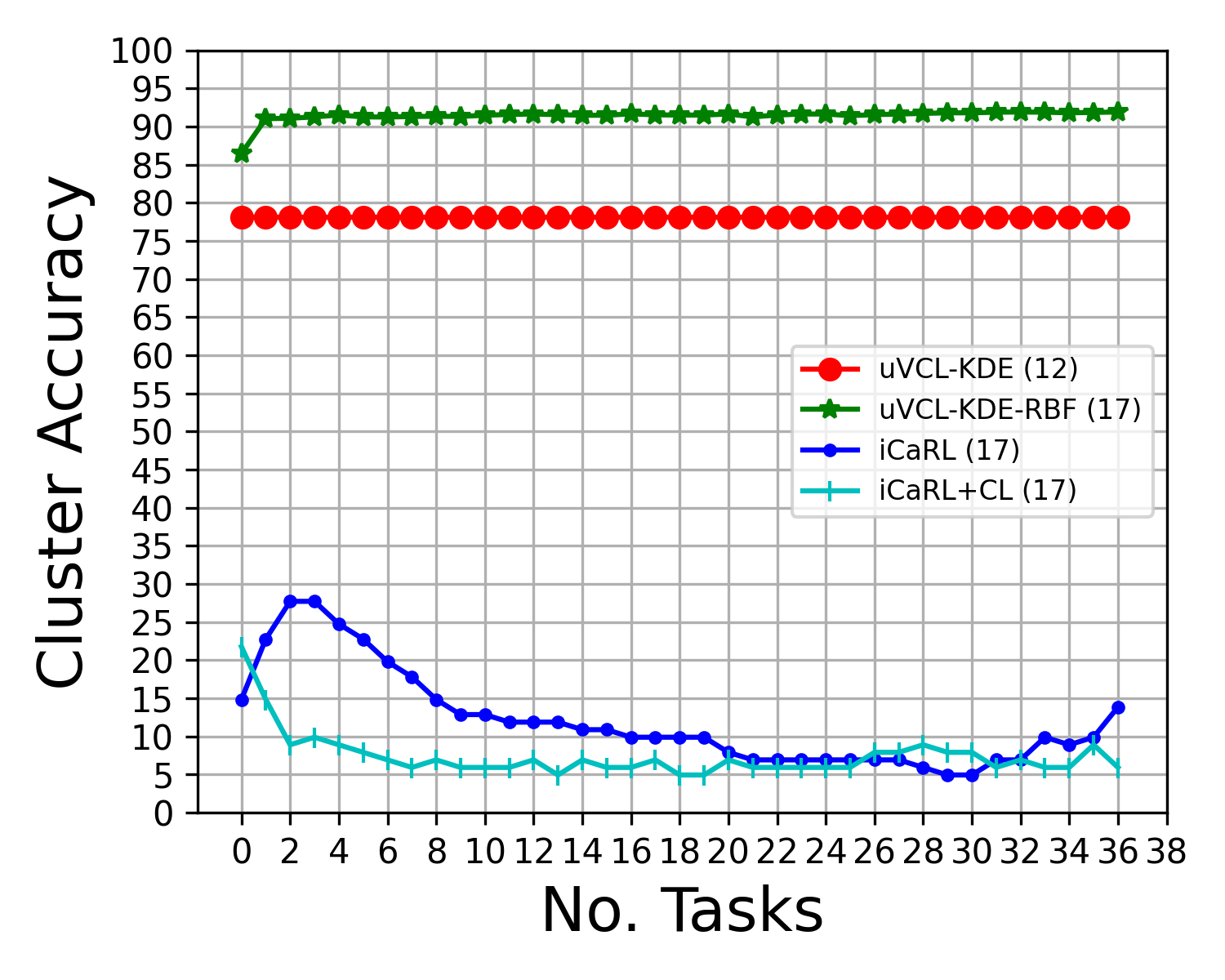} &
\includegraphics[width=0.3\textwidth]{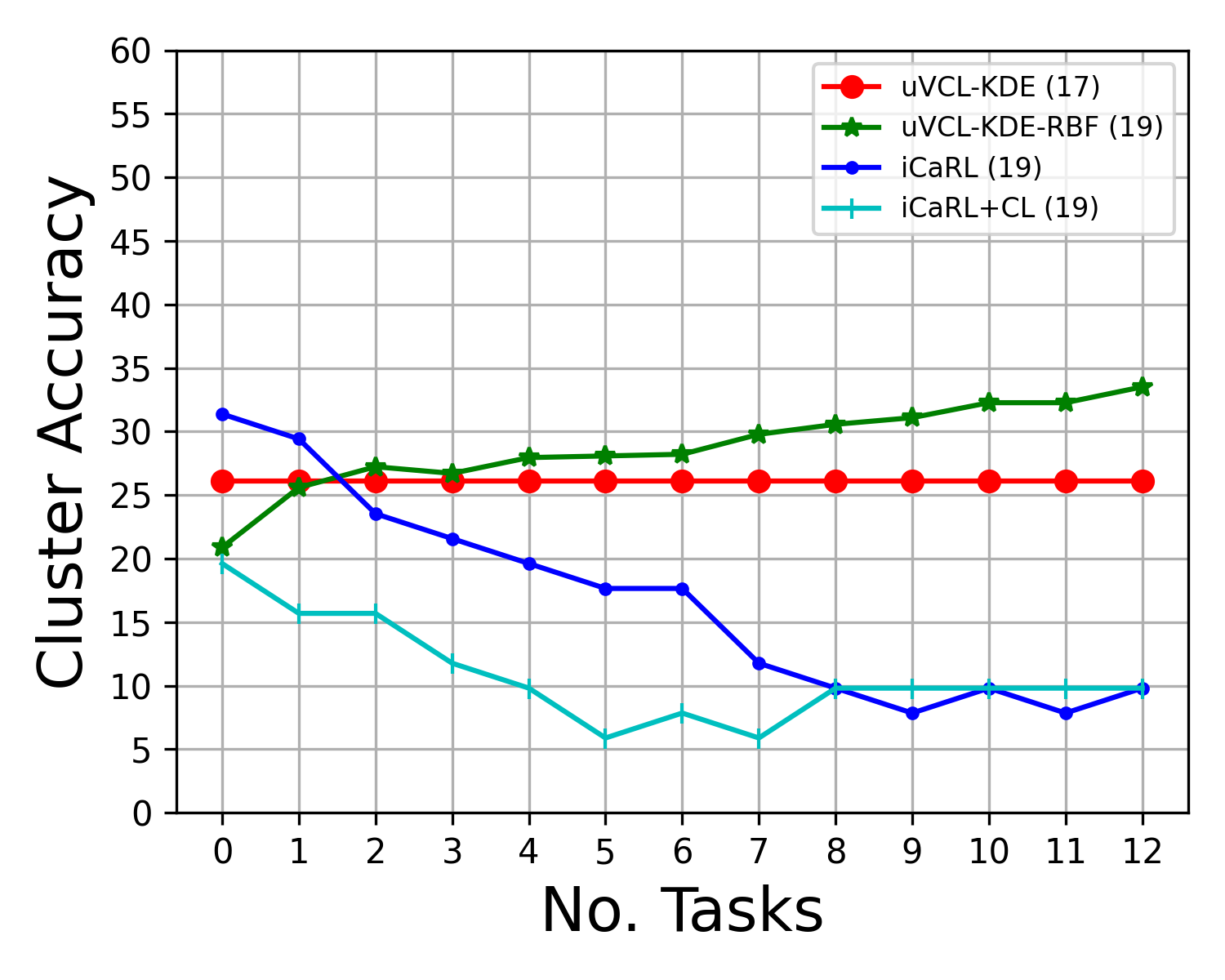} &
\includegraphics[width=0.3\textwidth]{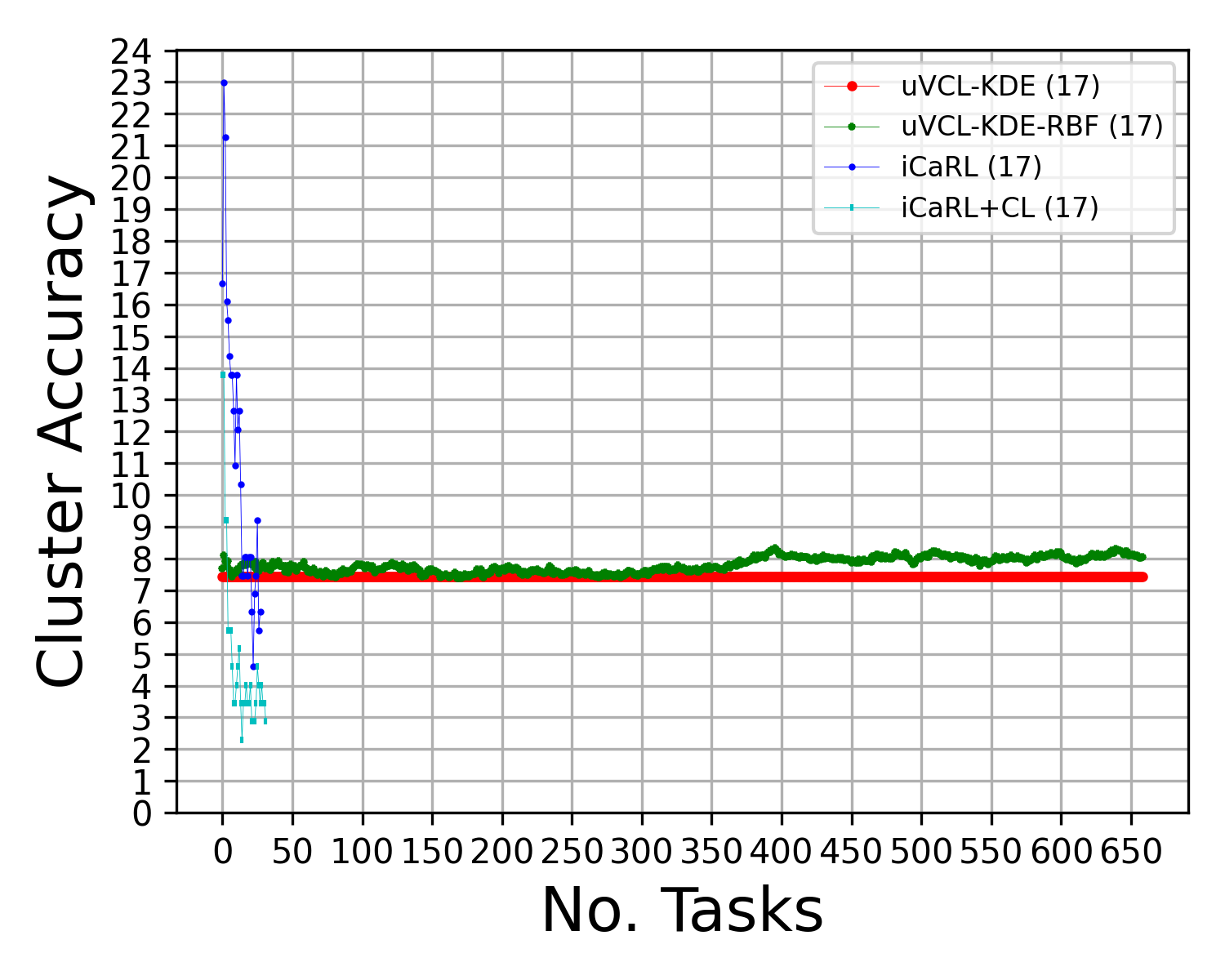}\\
{\small a) CAcc on UCF101.} & {\small b) CAcc on HMDB51.} & {\small c) CAcc on SSv2.} \\
\includegraphics[width=0.3\textwidth]{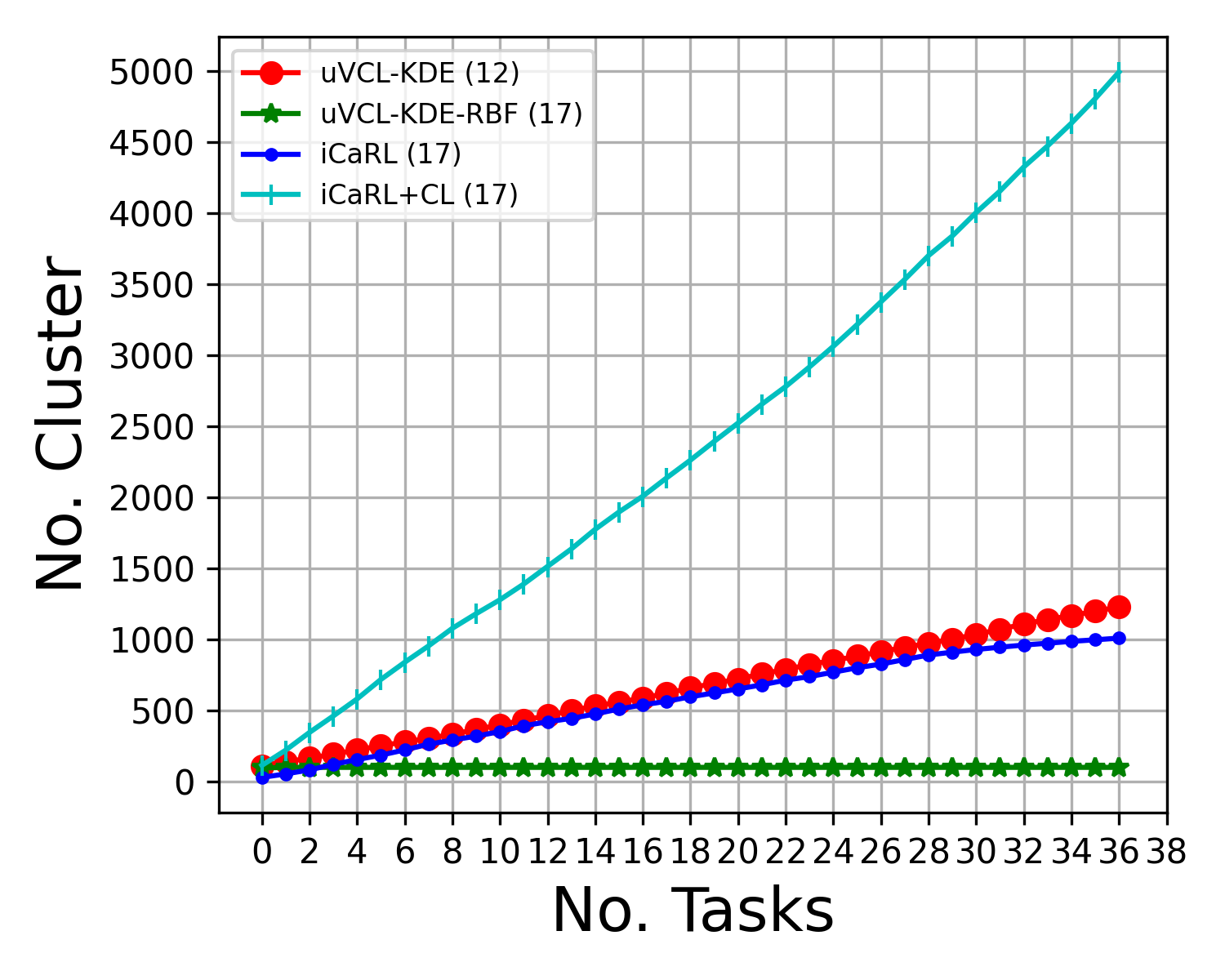}&
\includegraphics[width=0.3\textwidth]{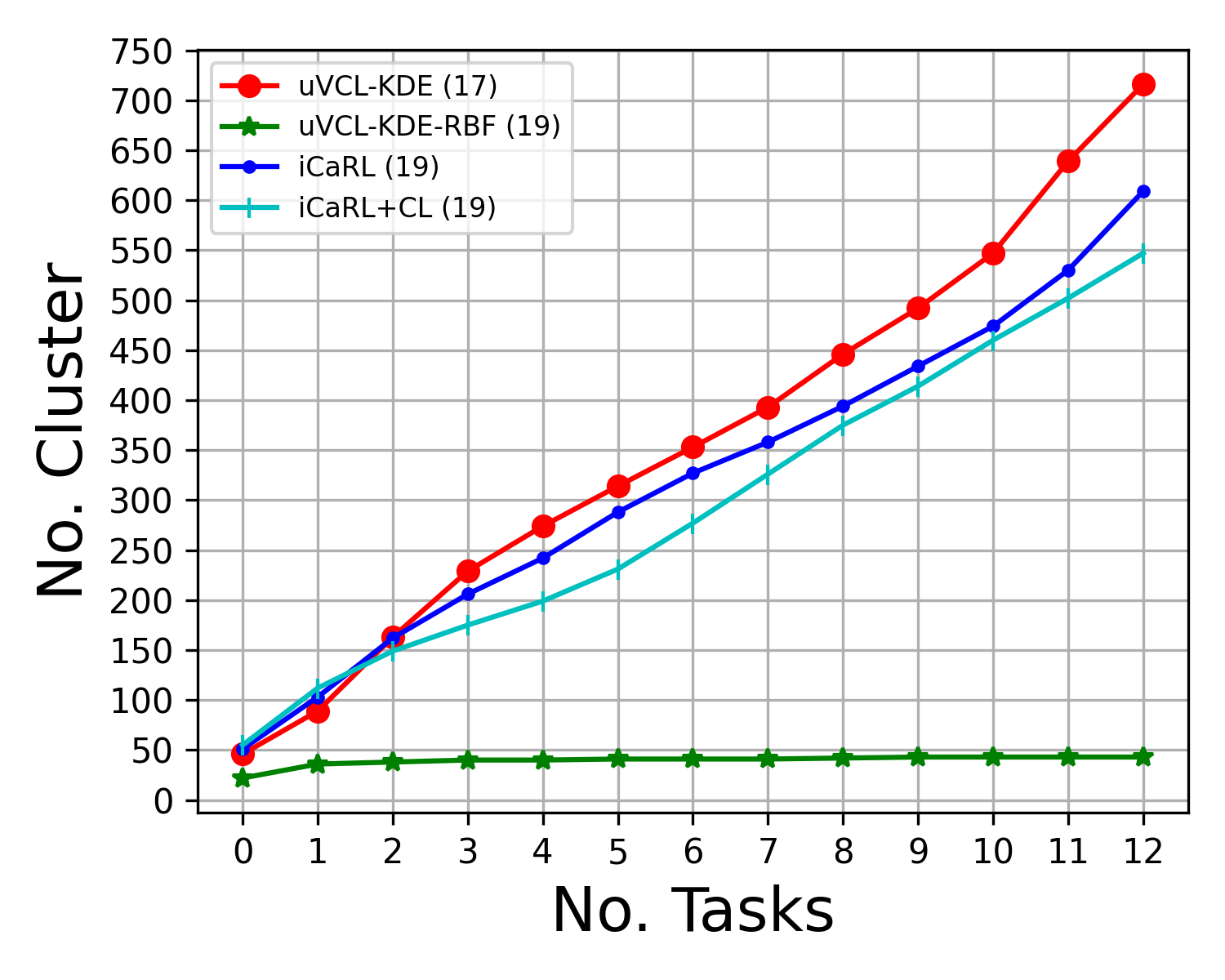}&
\includegraphics[width=0.3\textwidth]{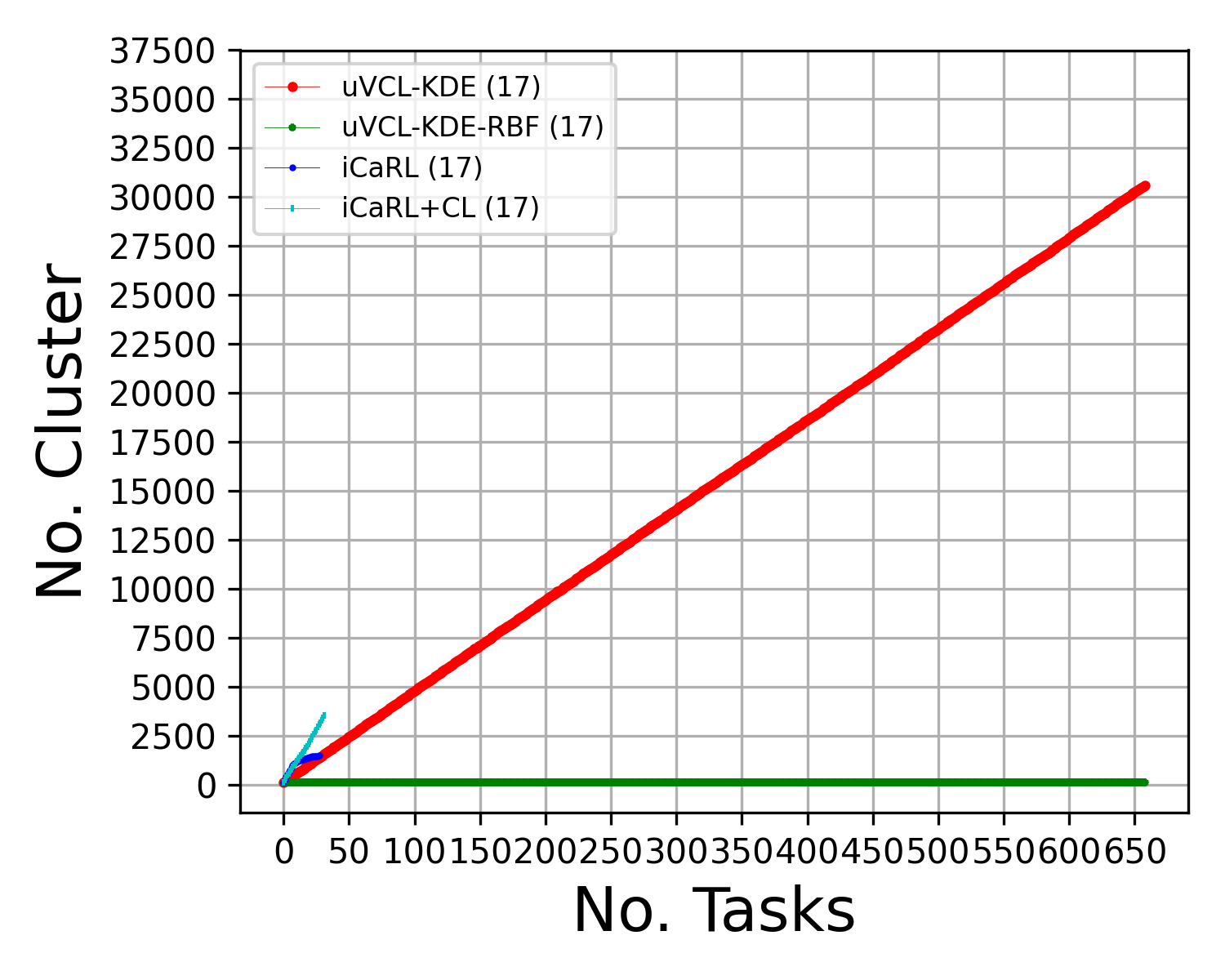}\\
{\small d) No. of clusters on UCF101.} & {\small e) No. of clusters on HMDB51.}  & {\small f) No. of clusters on SSv2.}
\end{tabular}
\end{center}
\caption{uVCL results on UCF101, HMDB51 and SSv2 considering the first fold data. Inside the brackets for each method we specify the bandwidth $h$ for the mean-shift clustering. }
\label{fig:main_result_learn_step}
\end{figure}

\subsection{uVCL-KDE results on UCF101, HMDB51 and SSv1}
\label{sec:uvcl_results}

We apply the clustering methodology proposed in this paper in Section~\ref{sec:deep_embedded_clustering} for uVCL-KDE and also its extension uVCL-KDE-RBF, described in Section~\ref{RBF-linear}, where the liniear units are randomly initialized.
We present the results on UCF101, HMDB51, and Something-Something V2 (SSv2) datasets, where we split the data into 13, 37 and 659 tasks, considering 256 videos from a random mixed of classes, at each task. For UCF101 and HMDB51 datasets, the results averaged across three different data splits are shown in Table~\ref{tab:sumary_kde}. In our experiments, we consider adjusting the bandwidth with various values for $h$ for the Mean-shift algorithm in order to find the best value. We report the average final number of clusters ($L_k$), average final cluster accuracy (CAcc) over three data splits for UCF101 and HMDB51, average cluster accuracy (ACAcc) over all learning tasks, backward forgetting (BWF), and forward forgetting (FWF). The best results are obtained for $\Theta_2 = 0.3$ in Eq.~\eqref{eq:theta_2}, and for the bandwith $h = 17$ for UCF101 and SSv2 whereas $h = 19$ in HMDB51. According to the results from Table~\ref{tab:sumary_kde}, uVCL-KDE-RBF achieves the best results for UCF101, HMDB51 and SSv2, by considering 100, 42 and 133 clusters, respectively. 

\begin{table}[ht!]
\begin{center}
\resizebox{\linewidth}{!}{
\begin{tabular}{lc|ccccc|ccccc|ccccc}
\hline
\multirow{ 3}{*}{Methods}   &   \multirow{ 3}{*}{ \makecell{ h }}   &  \multicolumn{5}{c|}{UCF101}  &  \multicolumn{5}{c|}{HMDB51}  &  \multicolumn{5}{c}{SSv2}        \\ \cline{3-17}   
&  &     \makecell{ Avg \\ ${L_k}$} &     \makecell{ Avg \\ ${CAcc_k}$} &   \makecell{  Avg \\ ACAcc} &   \makecell{ $BWF_k$ $\uparrow$ } &   \makecell{ $FWF_k$ $\downarrow$} &     \makecell{ Avg \\ ${L_k}$} &     \makecell{ Avg \\ ${CAcc_k}$} &   \makecell{  Avg \\ ACAcc} &   \makecell{ $BWF_k$ $\uparrow$ } &   \makecell{ $FWF_k$ $\downarrow$} &     \makecell{ Avg \\ ${L_k}$} &     \makecell{ Avg \\ ${CAcc_k}$} &   \makecell{  Avg \\ ACAcc} &   \makecell{ $BWF_k$ $\uparrow$ } &   \makecell{ $FWF_k$ $\downarrow$}         \\
\hline
\multirow{4}{*}{\makecell{ uVCL-KDE \\ } }           
             &           15 &             770 &                86.79 & 86.79  & 0.0 & 0.0 &             1,140 &                23.75 & 23.75  & 0.0 & 0.0 & 87,628 &	7.52	& 7.52	& 0.0	& 0.0 \\ 
             &           16 &             657 &                86.90 & 86.90  & 0.0 & 0.0 &             914 &                26.51 & 26.51  & 0.0 & 0.0 &             59,895 &                7.68 & 7.68  & 0.0 & 0.0 \\
             &           17 &             506 &                87.46 & 87.46  & 0.0 & 0.0 &             693 &                28.02 & 28.02  & 0.0 & 0.0 &             30,603 &                7.44 & 7.44  & 0.0 & 0.0 \\
             &           18 &             518 &                82.98 & 82.98  & 0.0 & 0.0 &             486 &                27.26 & 27.26  & 0.0 & 0.0 &             15,658 &                7.27 & 7.27  & 0.0 & 0.0\\
             \hline 
  \multirow{4}{*}{\makecell{ uVCL-KDE-RBF \\ }}            
             &           16 &                   101 &                92.80 & 92.57  & 0.22 & -0.11 & 117	& 22.45 &	21.34	& 1.08	& 0.23 &                  227 &                7.42 & 7.46  & 0.01 & 0.000
  \\ 
             &           17 &                   100 &                \textbf{93.45} & \textbf{93.01}  & \textbf{0.33} & \textbf{-0.15} &                  92 &                27.79 & 25.97  & 1.48 & 0.08 &                   133 &                \textbf{8.07} & 7.81  & 0.24 & -0.001 \\
             &           18 &                   93 &                88.27 & 88.05  & 0.16 & -0.12 &                   60 &                32.90 & 29.65  & 3.04 & -0.67 &                   79 &                8.02 & 7.64  & \textbf{0.32} & -0.002 \\
             &           19 &                   85	
 &                83.52	 & 83.31	  & 0.25	 & -0.12 &                   42 &                \textbf{34.05} & \textbf{29.81}  & \textbf{3.76} & \textbf{-0.99} &                   67 &                7.44 & 7.36  & 0.10 & -0.001 \\
             \hline
 \makecell{ iCaRL  \cite{iCaRL} \\ }                &          \multirow{4}{*}{ \makecell{ 17\\or\\19 } }   &                  727	 &                10.23	 & 12.34	  & -2.16	 & 0.01 & 647  &                  9.15	 &                14.78	 & -6.10	  & 1.58 & 1,486	&6.32	& \textbf{11.00}	&-4.85	&0.38
 \\
 \makecell{ iCaRL+CL \cite{vCLIMB} \\ }             &             &                   5,052	
 &                5.94	 & 7.15	  & -1.25	 & 0.27 & 542	 &                11.11	 & 10.76	  & 0.38	 & 0.38  & 3,553	&3.45	&4.92	&-1.53	&0.36

\\ 
 \makecell{ EWC  \cite{EWC} \\ }             &             &                   42	
 &                2.31	 & 1.92	  & 0.40	 & -0.04 & 13		 &                3.92		 &3.82		  & 0.11		 & -0.16 & 81	&1.15	&0.98	&0.18	&-0.02
  \\ 
 \makecell{ MAS  \cite{aljundi2018memory} \\ }             &             &                   37	
 &                11.55	 & 7.59	  & 4.07	 & -0.29 & 13	 &   4.58		 & 4.68		  & -0.11		 & -0.22 & 89	&5.17	&5.38	&-0.22	& \textbf{-0.16}
 \\ 
            \hline
\end{tabular}
}
\end{center}
\caption{Unsupervised video continual learning results for UCF101, HMDB51, and SSv2, where the results represent the average across three data splits. The results for the SSv2 dataset are provided only for the first 30 tasks.}
\label{tab:sumary_kde}
\end{table}

We investigate the progressive learning of the proposed KDE-based methodology, task by task. The cluster accuracy (CAcc) are provided in Fig.~\ref{fig:main_result_learn_step}-a, b, c, while the number of clusters considered according to increasing the number of tasks are provided in Fig.~\ref{fig:main_result_learn_step}-d, e, f, respectively, for the continual learning of UCF101, HMDB51, and SSv2, respectively. These results show that uVCL-KDE-RBF achieves better results than all other baselines considered as well as than uVCL-KDE. The proposed method is shown to maintain and improve its performance over the successive learning of the tasks under all evaluation metrics. Moreover, our uVCL-KDE-RBF model finds a number of clusters  which is close to the ground truth class number, assued to be the number of classes. In addition, the baseline experiment on the SSv2 dataset is conducted only for the first 30 tasks because they require significant memory and significant computation costs for training, with the result showing a trend to a dramatic reduction in performance from the very beginning on this challenging dataset. Furthermore, results for the Backword Forgetting (BWF) and the Forward Forgetting (FWF) for all datasets are provided and explained in the Appendix E from SM.

\subsection{Ablation study}
\label{sec:ablation}

{\bf  Changing the size of the memory buffer.}
We consider storing the features corresponding to $N=20$ videos for each cluster, similar to the supervised video class incremental study from \cite{vCLIMB}, 
where, unlike in our study, the class labels were known. 
Due to the inevitable variations in the size of each category, 
a small fixed memory size could lead to the loss of critical examples. 
To address this limitation, we consider a dynamic memory size for storing data, starting by keeping 10 examples per cluster and gradually expanding to 30 per cluster. The data stored is randomly selected from the data associated with each cluster. The results provided in Table \ref{tab:ucf101_hmdb51_kde_change_memory_buffer} show that by increasing the memory size to 30 samples per cluster, results in a better performance on UCF101, while a smaller buffer of 10 examples is more effective for HMDB51. 

\begin{table}[ht!]
\begin{center}
\resizebox{\linewidth}{!}{
\begin{tabular}{lcc|ccccc|ccccc}
\hline
 \multirow{ 3}{*}{\makecell{Methods}}   &   \multirow{ 3}{*}{ \makecell{ h }} & \multirow{ 3}{*}{\makecell{ Memory \\Size}}    &  \multicolumn{5}{c|}{UCF101} & \multicolumn{5}{c}{HMDB51}         \\ \cline{4-8} \cline{9-13}   
    &    &  & \makecell{ Avg \\ ${L_k}$} &     \makecell{ Avg \\ ${CAcc_k}$} &   \makecell{  Avg \\ ACAcc} &   \makecell{ $BWF_k$ $\downarrow$ } &   \makecell{ $FWF_k$ $\uparrow$} &     \makecell{ Avg \\ ${L_k}$} &     \makecell{ Avg \\ ${CAcc_k}$} &   \makecell{  Avg \\ ACAcc} &   \makecell{ $BWF_k$ $\downarrow$ } &   \makecell{ $FWF_k$ $\uparrow$}       \\
\hline
 \multirow{4}{*}{\makecell{uVCL-KDE-RBF}}  
            &           17 & \multirow{2}{*}{\makecell{10 examples/\\cluster}}   & 98	 & 91.51	 & 91.14	 & 0.44	 & -0.16 & 87	 & 27.44	 & 25.86	 & \textbf{0.27}	 & \textbf{0.21} \\
            &           19 &  & 85	 & 83.25	&82.60	 & 0.66	 & -0.17 & 46 & \textbf{34.14} & \textbf{30.11} & 3.82 & -0.95  \\ \cline{2-13} 
            &           17 & \multirow{2}{*}{\makecell{30 examples/\\cluster}}   & 97	 & \textbf{92.65}	 & \textbf{92.33}	 & 0.31	 & -0.13 & 82	
 & 27.64	 & 26.54 & 1.00	
 & 0.03 \\
            &           19 &  & 86	 & 84.39	&84.20	 & \textbf{0.22}	 & \textbf{-0.08} & 40	 & 31.04	 & 26.98	 & 4.28	 & -0.72 \\

\hline
\end{tabular}
}
\end{center}
\caption{The performance on UCF101 and HMDB51 when we change the number of data stored in the memory buffers for each cluster, at 10 and 30 examples per cluster. }
\label{tab:ucf101_hmdb51_kde_change_memory_buffer}
\end{table}

\noindent
{\bf Changing the thresholds $\Theta_1$ and $\Theta_2$.} 
For novelty detector threshold $\Theta_1$ for UVCL-KDE in Eq~\eqref{eq:Theta1}, We use $\Theta_1 = 16.53$, $\Theta_1 =  16.92$, $\Theta_1 =  15.98$, for UCF101, HMDB51, and SSv2, respectively. Moreover, we vary the novelty detector threshold  in Eq.~\eqref{eq:theta_2}, by increasing from $\Theta_2$ which controls the confidence for creating new clusters and assigning them pseudo-labels when learning new tasks. A smaller $\Theta_2$  allows more clusters to be created, while higher thresholds are more selective, potentially avoiding incorrect cluster assignments. We experiment on UCF101 and HMDB51 datasets, and the results are shown in Table \ref{tab:ucf101_hmdb51_kde_change_threshold}. We conclude that a small threshold at $\Theta_2 = 0.3$ performs the best, resulting in semantically meaningful clusters. We provide the computation costs and the number of parameters required by each model in Appendix - F from the SM.

\begin{table}[ht!]
\begin{center}
\resizebox{\linewidth}{!}{
\begin{tabular}{lcc|ccccc|ccccc}
\hline
 \multirow{ 3}{*}{\makecell{Methods}}   &   \multirow{ 3}{*}{ \makecell{ h }} & \multirow{ 3}{*}{\makecell{ $\Theta_2$ }}    &  \multicolumn{5}{c|}{UCF101} & \multicolumn{5}{c}{HMDB51}         \\ \cline{4-8} \cline{9-13}   
    &    &  & \makecell{ Avg \\ ${L_k}$} &     \makecell{ Avg \\ ${CAcc_k}$} &   \makecell{  Avg \\ ACAcc} &   \makecell{ $BWF_k$ $\uparrow$ } &   \makecell{ $FWF_k$ $\downarrow$} &     \makecell{ Avg \\ ${L_k}$} &     \makecell{ Avg \\ ${CAcc_k}$} &   \makecell{  Avg \\ ACAcc} &   \makecell{ $BWF_k$ $\uparrow$ } &   \makecell{ $FWF_k$ $\downarrow$}       \\
\hline
 \multirow{4}{*}{\makecell{uVCL-KDE-RBF}}  
            &           \multirow{2}{*}{\makecell{17}} & 0.7   & 118	
 & \textbf{92.69}	 & \textbf{92.47}	 & \textbf{0.26}	 & -0.12 & 331	& 19.44	 & 21.25	 & -1.54	 & 0.75 \\
            &            & 1.0   & 1,321	 & 57.53	 & 59.32	 & -1.78	 & 0.88 & 748	
 & 10.99	 & 13.99	 & -4.03	 & 1.50 \\ \cline{2-13}
            &           \multirow{2}{*}{\makecell{19}} &  0.7 & 106	 & 85.72	& 85.51	 & 0.23	 & \textbf{-0.18} & 192	
 & \textbf{28.10}	 & \textbf{28.29}	 & \textbf{-0.14}	 & \textbf{-0.55} \\ 
            &            & 1.0 & 1,180	 & 59.54	& 60.43	 & -1.46	 & 0.58 & 384	
 & 17.67	 & 21.13	 & -3.89	 & 0.30 \\
\hline
\end{tabular}
}
\end{center}
\caption{The performance on UCF101 and HMDB51 when changing the value of the threshold $\Theta_2 \in \{0.7, 1.0\}$ for defining new clusters in Eq.~\eqref{eq:theta_2}. We store $N = 20$ examples/cluster.}
\label{tab:ucf101_hmdb51_kde_change_threshold}
\end{table}

\section{Conclusion and future work}
\label{sec:conclusion}

In this paper, we propose a realistic yet effective framework for the Unsupervised Video Continual Learning (uVCL), which relies on dynamic kernel density estimation (KDE) representations for the features extracted by video transformers.
A number of clusters is built and managed dynamically. We propose two different approaches, one based on the mean-shift algorithm for representing KDE and extracting clusters of video data while the other uses a linear layer on top of the clusters as in the Radial Basis Function (RBF) networks, and it is named uVCL-KDE-RBF. The key to sustaining the performance is to use memory buffers, storing the video features of some data associated with each cluster. Such stored data is then used again when new tasks are introduced, ensuring that the model can recall prior information and mitigate catastrophic forgetting. Our experiments highlight that our proposed methods not only that it reduces computation requirements and training time but it also effectively preserves past knowledge balancing stability and plasticity in the unsupervised video continual learning. In future work, we will employ a dynamic novelty detector criterion for deciding when to learn new information and define new clusters. 

\clearpage
\bibliography{egbib}

\setcounter{section}{0}

\section{Appendix A - Memory management}

As we have no initial information on the number of clusters, we consider instead fixing the maximum memory size as it is used in methods such as iCaRL \cite{iCaRL} and iCaRL+CL \cite{vCLIMB}. In order to reduce the memory requirements we store the embedded features instead of the real video.  Thus, we proposed to use First-In First-Out (FIFO) for memory management, when associating data with a specific peak, representing one of the clusters considered. So the earliest samples associated with a cluster are removed when new samples are associated with the memory buffer ${\cal M}_i$ for the cluster $i$. This approach controls the memory requirements for the proposed UCL methodology.

\section{Appendix B -Datasets and Tasks}

We evaluate our proposed approach using three standard video action recognition datasets by ignoring the labels of the videos in order to follow an unsupervised learning setting. The UCF101 \cite{UCF101} dataset contains 13,320 videos from 101 classes. The HMDB51 \cite{HMDB51} dataset consists of 6,766 videos across 51 action classes. Both are three predefined splits for training and testing. The Something-Something V2 \cite{SSV2} dataset is a large-scale dataset consisting of more complex videos, with 220,847 videos from 174 action classes. We divided the training data into a sequence of tasks, using 256 examples per task with a random mixed class for continual learning. The UCF101 will contain 37 tasks, HMDB51 will contain 13 tasks, and SSv2 will contain 659 tasks. More information is described in Table \ref{tab:data-statistics}.

\begin{table}[ht!]
\centering

\resizebox{\linewidth}{!}{%
\begin{tabular}{lcccc}
\hline
Datasets  & Tasks &   Train & Test  & Video Training  \\  
          &       &          &      &  Data Size/Task \\ \hline
HMDB51 Fold-1,2,3     & 13    & 3,570       &  1,530    & 256     \\ \hline
UCF101 Fold-1          & 37    & 9,537          & 3,783     & \multirow{3}{*}{256}   \\
UCF101 Fold-2         &  37                      & 9,586          & 3,734     &      \\
UCF101 Fold-3         &  37                      & 9,624          & 3,696     &      \\ \hline
Something-Something V2  & 659                     & 168,913          & 27,157     & 256  \\ \hline
\end{tabular}%
}
\caption{The characteristics of the videos used for the unsupervised continual learning.}
\label{tab:data-statistics}
\end{table}

\section{Appendix C - Baselines used in the experiments}

We compare our proposed approach following the adaptation of well-known existing class-incremental methods to the unsupervised continual learning, considering the same data splits and equivalent memory size for a fair comparison with our methodology. We re-implement and evaluate four well-known supervised continual learning methods for unsupervised continual learning methods. Two replay-based baselines with memory storage are included with iCaRL \cite{iCaRL} and iCaRL+CL (with and without consistency loss) \cite{vCLIMB}, and two regularisation-based without memory storage, including MAS \cite{aljundi2018memory} and EWC \cite{EWC}. These adaptations from the open source code from vCLIMB \cite{vCLIMB} are based on the Temporal Segment Network (TSN) \cite{TSN} with a ResNet-34 backbone. The temporal data augmentation, as proposed in \cite{TSN}, is also applied. 
The mean-shift clustering with a Gaussian kernel is used to assign pseudo-labels. When considering the baselines, we preserve the same ratio of videos per class as in \cite{iCaRL,vCLIMB}, which is 20 videos per cluster, where we assume that each task introduces new 128 clusters. Therefore, the baseline model defines a memory that can save the information corresponding to 1,600,000, 8,320, and 94,720 videos for Something-to-Something V2, HMDB51, and UCF101, respectively. This assumption allows the baseline can keep 100\% of the training data in total. Which leads to huge computational cost and memory consumption.

\section{Appendix D - Evaluation Metrics}

For the evaluation, we adapt the protocol used in the unsupervised settings from \cite{caron2018deep,van2020scan}, evaluating the cluster accuracy (CAcc), used for the unsupervised continual learning for images \cite{he2021unsupervised}. 
First, we employ the Hungarian matching algorithm \cite{kuhn1955hungarian} to associate each pseudo-label of a cluster with a ground truth label, where the video labels are considered only for testing and not for training. We then compare the ground truth label of the testing sample with that associated with its corresponding cluster. We calculate the cluster accuracy according to the ratio $M_c/M$ between the number of correctly classified data $M_c$ and that of testing data $M$. In this work, CAcc is used to evaluate the model’s ability to provide semantic meaningful clusters. Moreover, we evaluate the average unsupervised continual learning accuracy over all the training tasks, including the final task (ACAcc) \cite{lopez2017gradient, vCLIMB}, as:
\begin{equation}
  ACAcc = \frac{1}{k} \sum_{j=1}^{k} (CAcc_{j}),
\label{eq:ACACC}
\end{equation}
To measure the influence of the learned task $k$ in the performance of future tasks we evaluate the Forward Forgetting (FWF) \cite{lopez2017gradient}~:
\begin{equation}
  FWF_k = \frac{1}{T_k - 1} \sum_{j=2}^{T_k} (CAcc_{j-1} - CAcc_{j}),
\label{eq:FWF}
\end{equation}
where $T_k$ is the number of learned tasks after learning the task $k$, and $CAcc_{j-1}$ and $CAcc_{j}$ represents the cluster accuracy on the task $j-1$ and task $j$, respectively. The positive Forward Forgetting when learning task $k$ decreases the performance on the previous task $k-1$. On the other hand, the negative Forward Forgetting when learning task $k$ increases the performance on the previous task $k-1$. A large positive Forward Forgetting is also known as catastrophic forgetting.

\begin{figure}
\begin{center}
\begin{tabular}{ccc}
\includegraphics[width=0.3\textwidth]{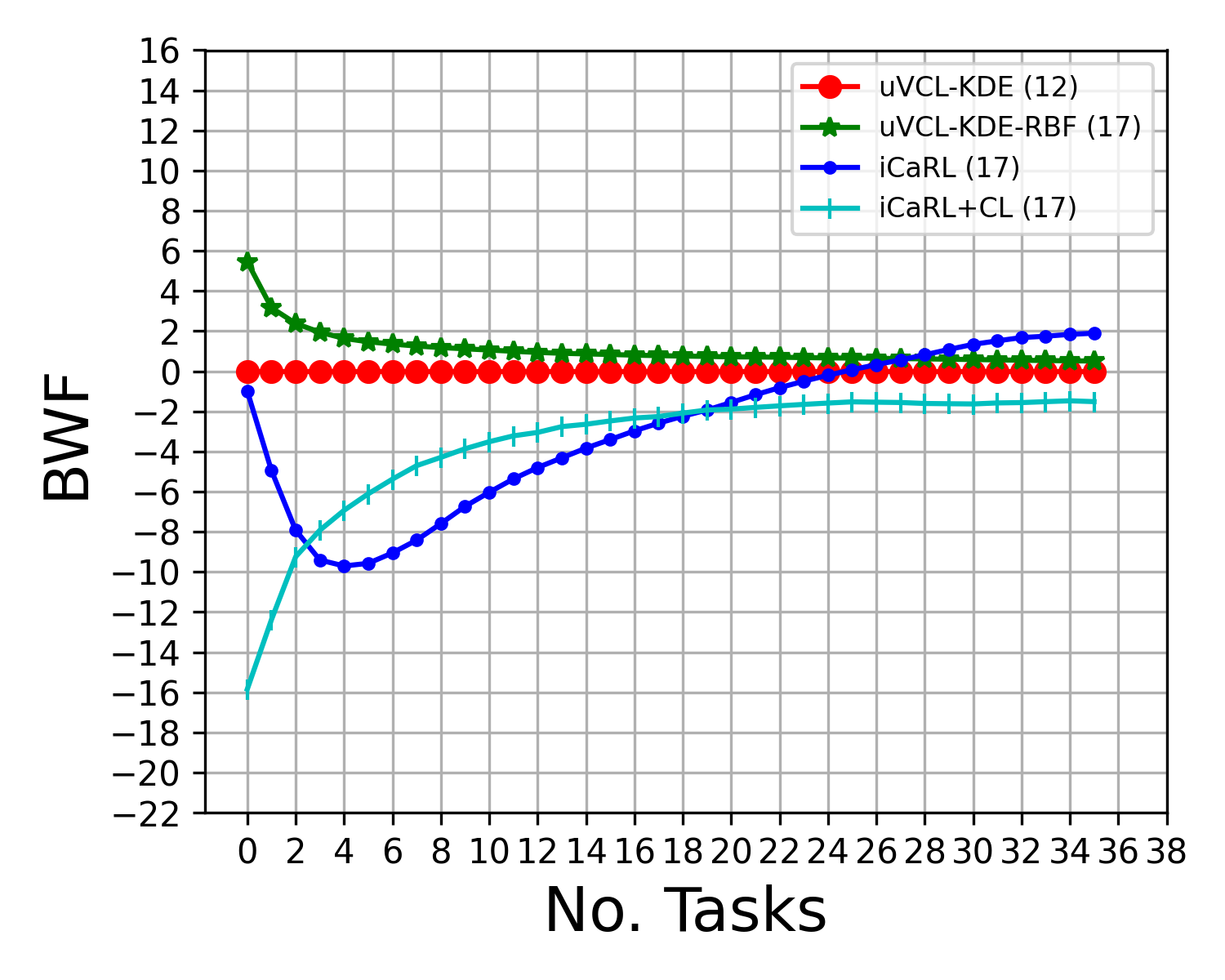}&
\includegraphics[width=0.3\textwidth]{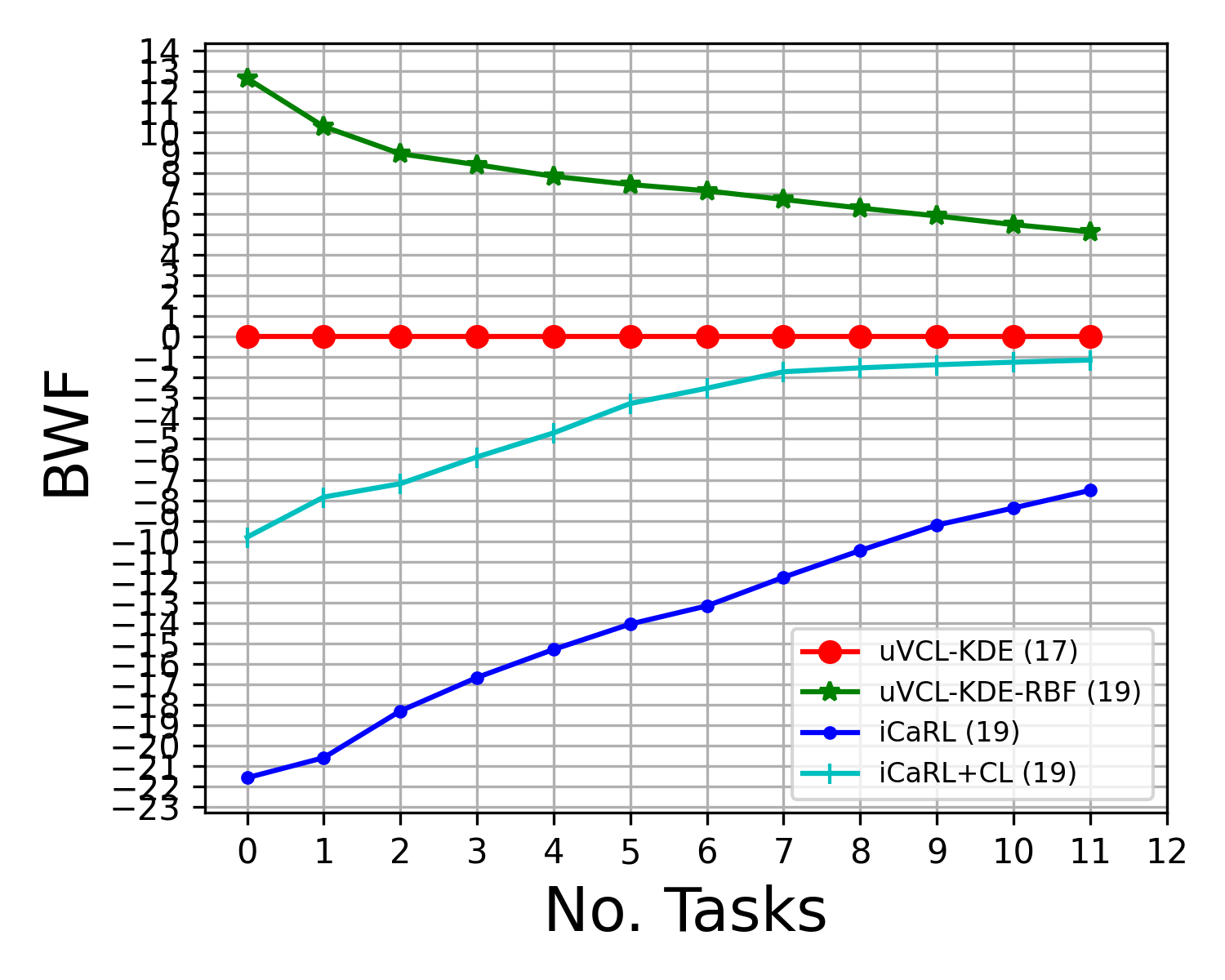}&
\includegraphics[width=0.3\textwidth]{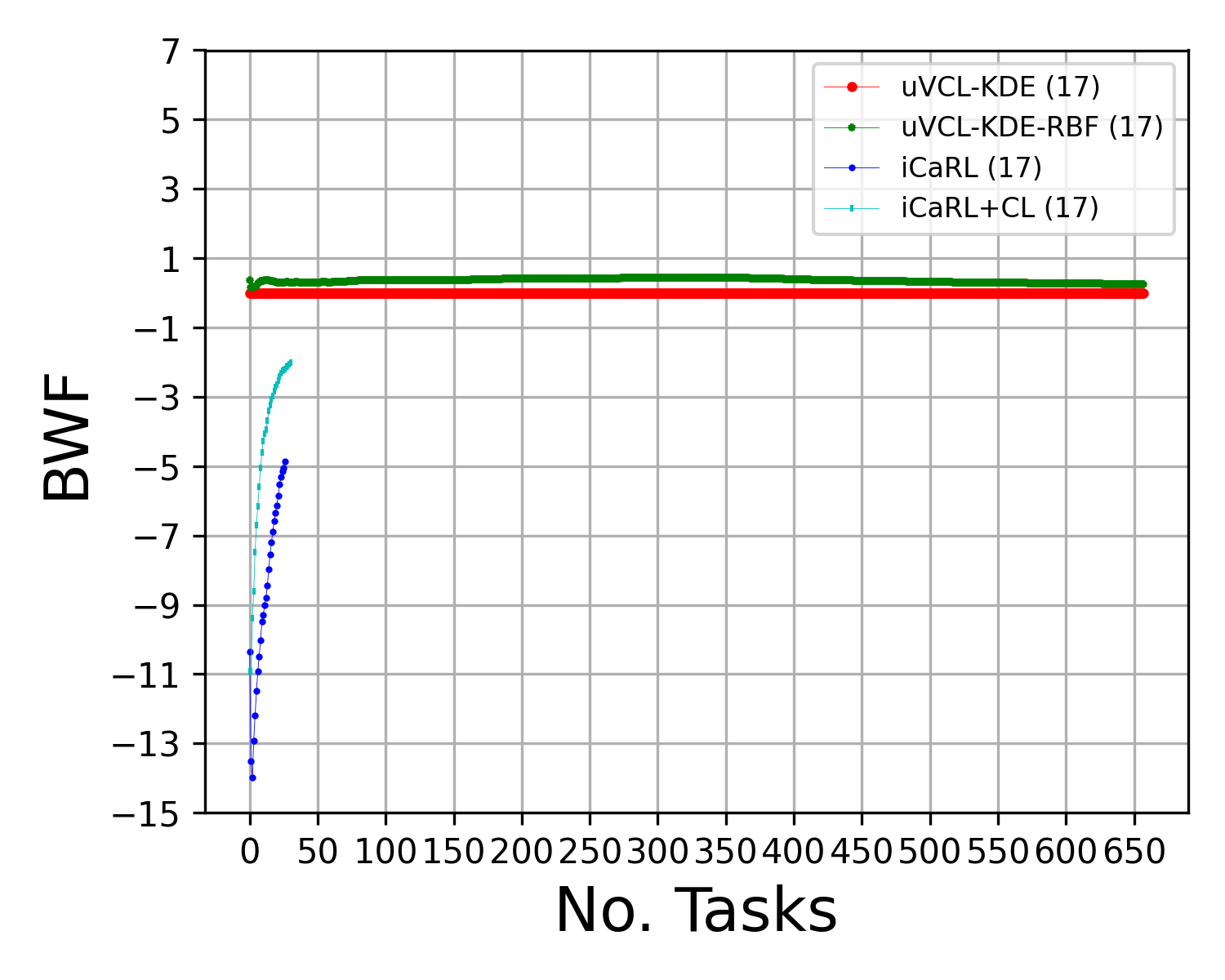}\\
{\small a) UCF101 BWF $\uparrow$} & {\small b) HMDB51 BWF$\uparrow$} & {\small c) SSv2 BWF$\uparrow$} \\
\includegraphics[width=0.3\textwidth]{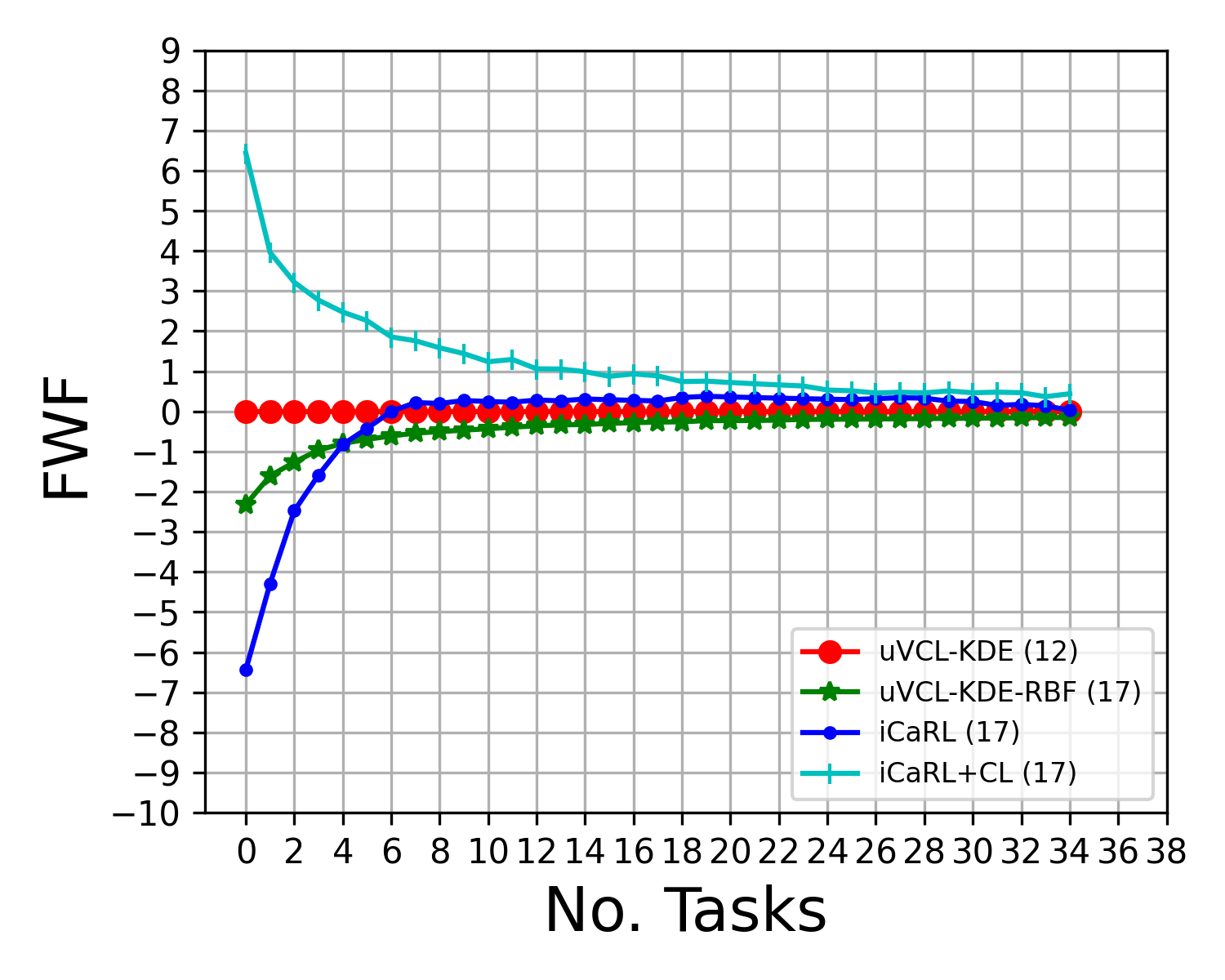}&
\includegraphics[width=0.3\textwidth]{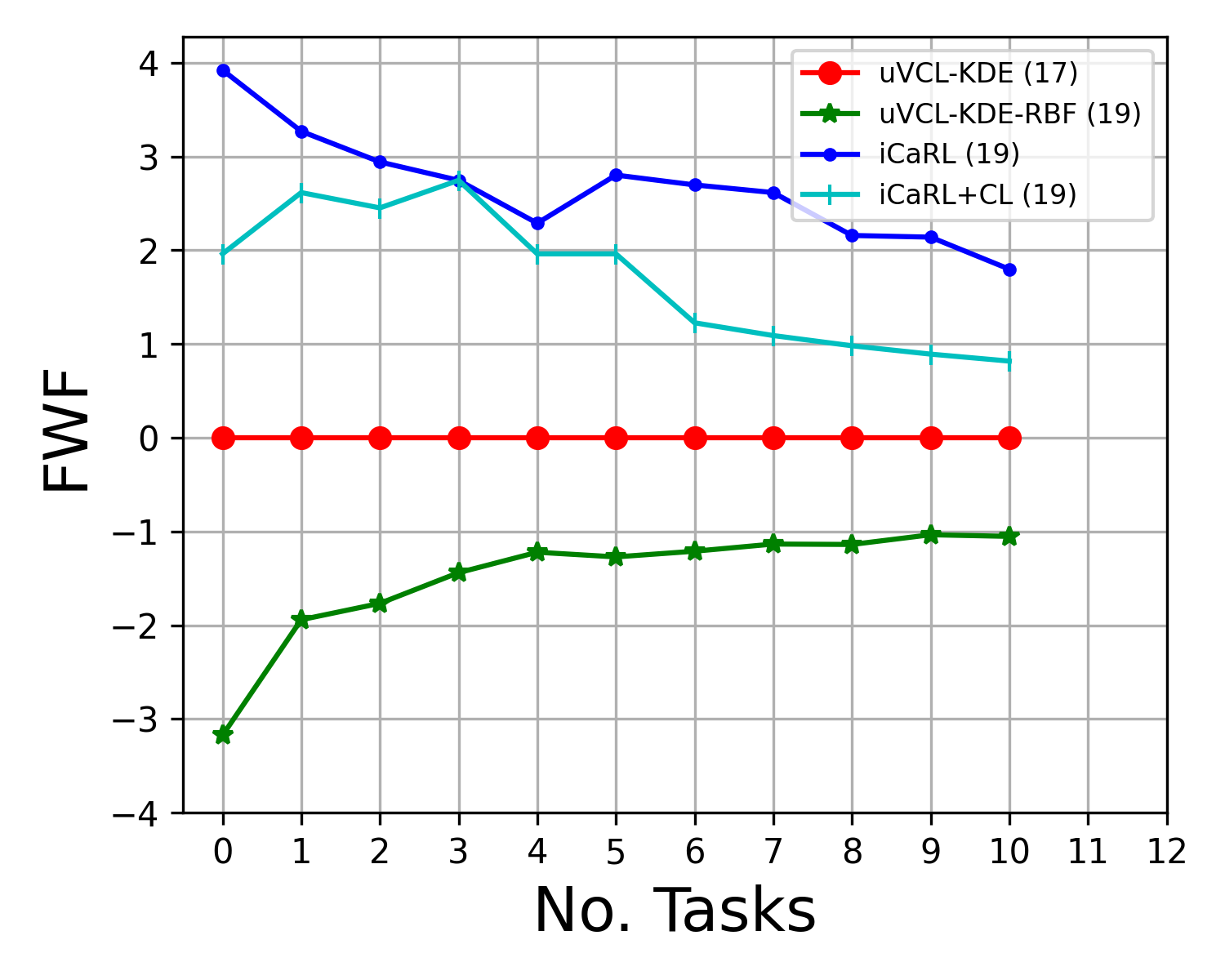}&
\includegraphics[width=0.3\textwidth]{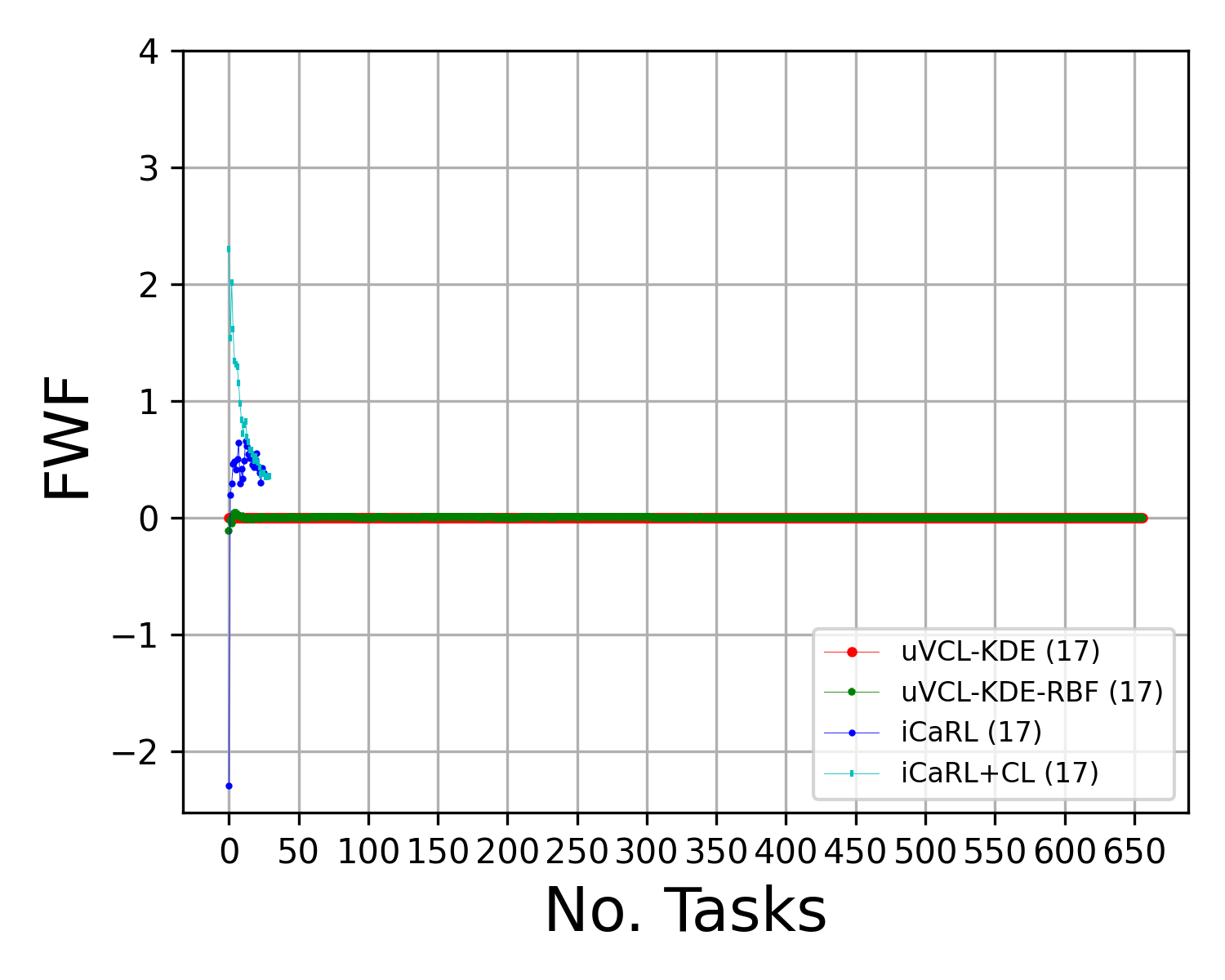}\\
{\small d) UCF101 FWF$\downarrow$} & {\small e) HMDB51 FWF$\downarrow$} &{\small f) SSv2 FWF$\downarrow$} \\
\end{tabular}
\end{center}
\caption{The evaluation of the Backword Forgetting (BWF) from Eq.~\eqref{eq:BWF} and the Forward Forgetting (FWF)  Eq.~\eqref{eq:FWF} results for the UVCL on UCF101, HMDB51 and SSv2 datasets, considering the first fold data. }
        \label{fig:main_result_learn_step_bwf_fwf}
\end{figure}

Moreover, to measure the influence of the learned task $k$ in the performance of the previous task, we also monitor Backword Forgetting (BWF) \cite{lopez2017gradient, vCLIMB}, as:
\begin{equation}
  BWF_k = \frac{1}{T_k - 1} \sum_{j=1}^{T_k - 1} ( CAcc_{T_{k}} - CAcc_{j}),
\label{eq:BWF}
\end{equation}
where $T_k$ is the number of learned tasks after learning the task $k$, and $CAcc_{j}$ and $CAcc_{T_{k}}$ represents the cluster accuracy on the task $j$ and task $T_k$, respectively. The positive backwards forgetting when learning task $T_k$ increases the performance on preceding task $j$. The negative backwards forgetting when learning task $T_k$ decreases the performance on preceding task $j$. A large negative backward forgetting is also known as catastrophic forgetting.

\section{Appendix E - Experimental Results}

In Figure.~\ref{fig:main_result_learn_step_bwf_fwf}, we provide some additional experimental results for the proposed Unsupervised Video Continual Learning. In Figure.~\ref{fig:main_result_learn_step_bwf_fwf}-a, b, c, we provide the Backword Forgetting (BWF) from Eq.~\eqref{eq:BWF} when uVCL is applied on UCF101, HMDB51 and SSv2 datasets, respectively. Meanwhile, in  Figure.~\ref{fig:main_result_learn_step_bwf_fwf}-d, e and f we provide the Forward Forgetting (FWF)  Eq.~\eqref{eq:FWF} results for the UVCL on UCF101, HMDB51 and SSv2 datasets, respectively. Inside the brackets for each method, we specify the bandwidth $h$ for clustering in each task. The number of features memorized in each buffer is $N=20$ examples per cluster. The novelty threshold is set to $\Theta_2=0.3$ for uVCL-KDE-RBF. The results show that our proposed method can perform the best against the catastrophic forgetting problem. 

\begin{figure}
\begin{center}
\begin{tabular}{ccc}
\includegraphics[width=0.3\textwidth]{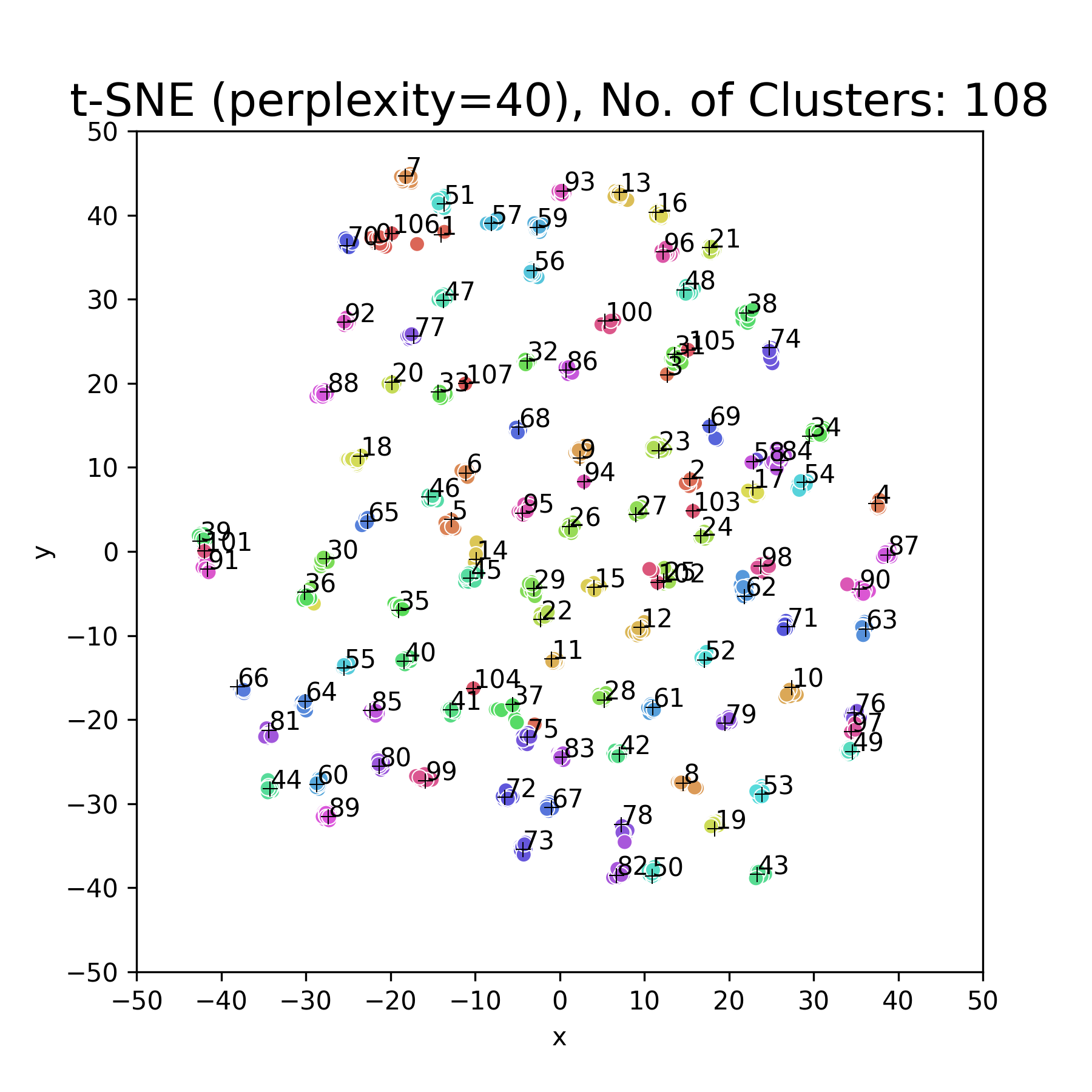}&
\includegraphics[width=0.3\textwidth]{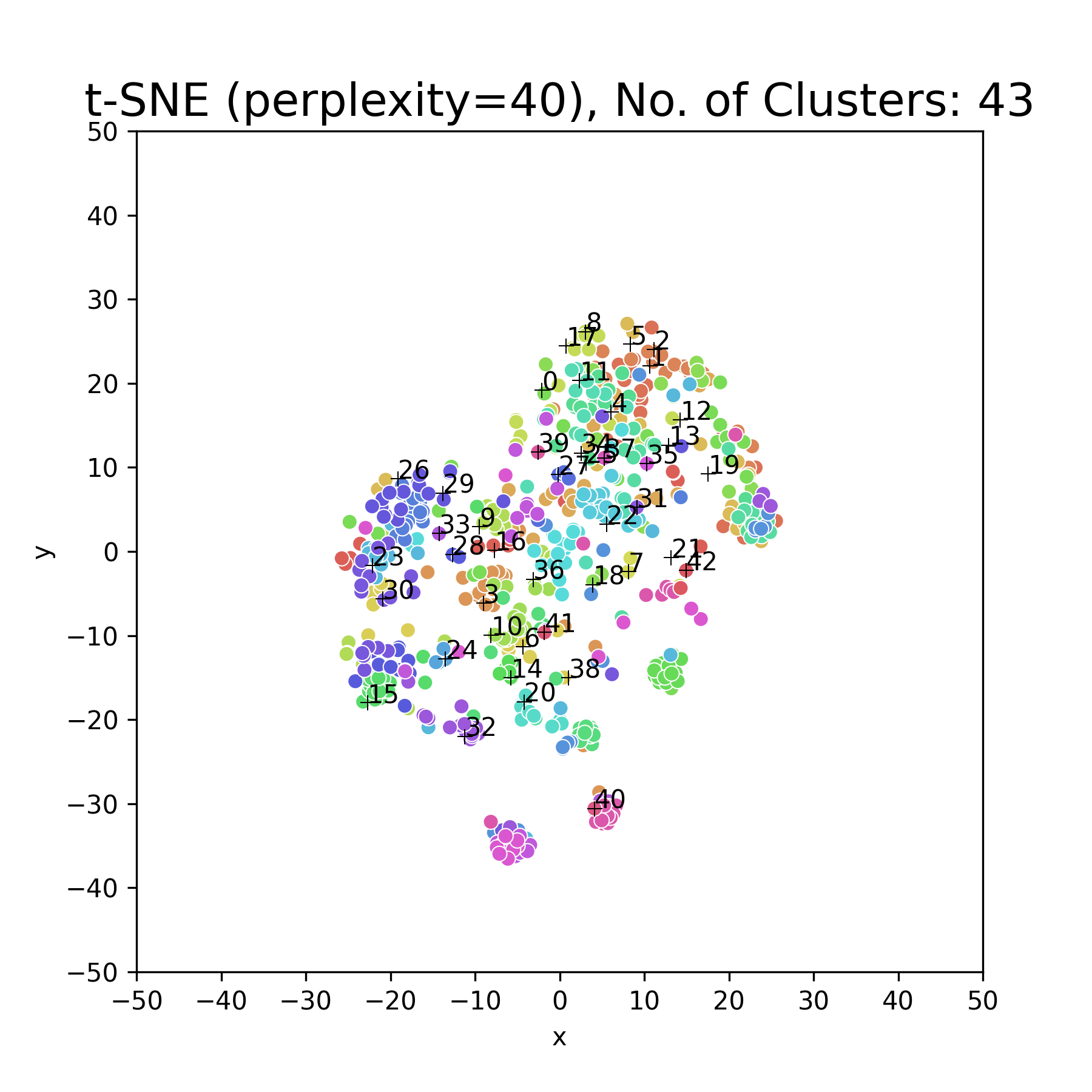}&
\includegraphics[width=0.3\textwidth]{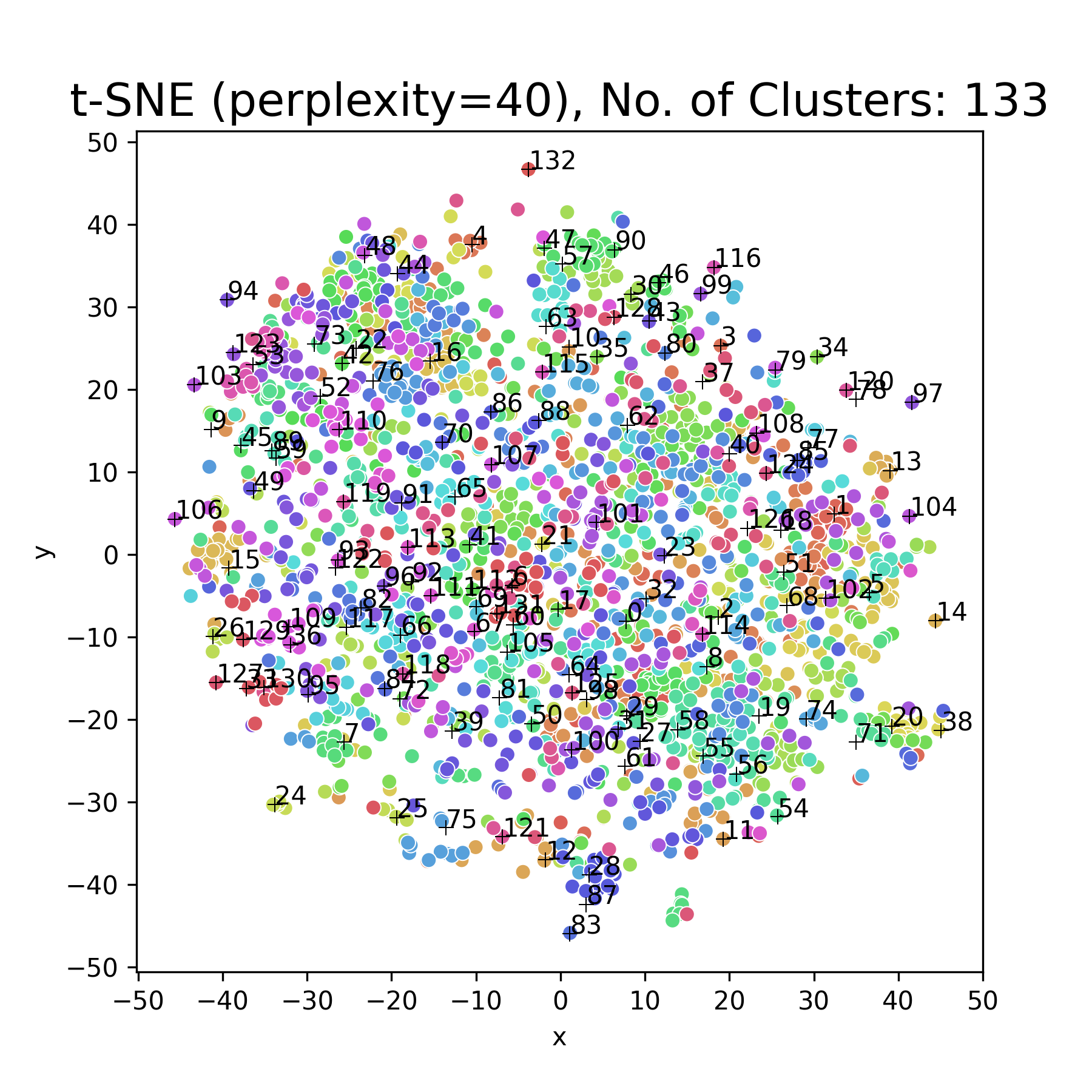}\\
{\small a) UCF101} & {\small b) HMDB51} & {\small c) SSv2} \\
\end{tabular}
\end{center}
\caption{We visualize the latent space stored in the memory for each cluster after learning all tasks by using t-SNE for feature reduction to 2-Dimensions with a perplexity of 40. This figure is best viewed in colour, $+$ represents the cluster centre, and a number represents the cluster ID. }
\label{fig:main_result_cluster}
\end{figure}

For visualisation of cluster distribution, we use t-SNE \cite{maaten2008visualizing} as inspired by \cite{xie2016unsupervised} applied to the embedded feature from the memory buffer, where the perplexity is set at 40. The result is shown in Figure \ref{fig:main_result_cluster}. It is clear that on UCF101, the clusters are significantly well separated as shown in Figure \ref{fig:main_result_cluster}(a). For HMDB51, as shown in Figure \ref{fig:main_result_cluster}(b), some clusters are well separated, whereas the less are close to the other cluster. For the SSv2 with a more complicated dataset, as shown in Figure \ref{fig:main_result_cluster}(c), the clusters are not well separated.

\section{Appendix F - The analysis of the computation cost and the number of parameters }

The computation complexity is essential to be considered when deploying the model on resource constrained systems. In Table~\ref{tab:compational_cost} we evaluate the number of parameters and the computational cost in the unsupervised continual learning of UCF101, HMDB51, and SSv2 for the proposed uVCIL-KDE and uVCIL-KDE-RBF as well as for other methods, such as iCaRL \cite{iCaRL} and iCaRL+CL \cite{vCLIMB}. The trainable parameters in uVCIL-KDE are computed by considering the number of clusters as $L_K \times |{\bf x}_{k,i}|$, where $L_K$ is the number of clusters
created until task $k$, and $|{\bf x}_{k,i}| = 1,024$ is the 
number of feature dimensions extracted by the unsupervised video autoencoder. For uVCIL-KDE we observe that the computational cost increases steadily with the number of clusters. The uVCIL-KDE-RBF uses a neural network for learning, where the number of trainable parameters increases slightly with 
the linear neural network built on top of the clusters. The computational cost remains relatively constant, with only a slight increase due to the complexity of the network, regardless of the number of clusters. According to Table~\ref{tab:compational_cost}, when comparing to other baselines, our proposed approach uses the least trainable parameters and the least training time. This means our proposed approach can learn faster than any other baseline. Especially on the SSv2 dataset, we found that the baseline method dramatically longer training time than our proposed approach without success in learning the task. Where our uVCL-KDE-RBF can learn 659 tasks in roughly 1 day and 43 minutes.

\begin{table}[ht!]
\begin{center}
\resizebox{\linewidth}{!}{
\begin{tabular}{lc|c|c|c|c|c|c|}
\hline
\multirow{3}{*}{\makecell{Methods}}  &  \multirow{3}{*}{\makecell{Feature \\Extractors\\ Parameters}} &  \multicolumn{2}{c|}{UCF101} & \multicolumn{2}{c|}{HMDB51} & \multicolumn{2}{c|}{SSv2} \\ \cline{3-8}     
      & & \multicolumn{1}{c|}{\makecell{Trainable\\ Parameters}} & \multicolumn{1}{c|}{\makecell{Training \\ Time (37 tasks) }} &  \multicolumn{1}{c|}{\makecell{Trainable\\ Parameters}} & \multicolumn{1}{c|}{\makecell{Training \\ Time (13 tasks) }}&  \multicolumn{1}{c|}{\makecell{Trainable\\ Paramameters}}& \multicolumn{1}{c|}{\makecell{Training \\ Time (659 tasks) }} \\
\hline
\multirow{1}{*}{\makecell{ uVCIL-KDE}}                  &  \multirow{2}{*}{\makecell{ 30.3M \\(VideoMAEv2 \cite{wang2023videomae})}}    &      518.14K & 0d 01h 23m 58s & 338.94K & 0d 00h 29m 10s & 61.33M & 1w 3d 06h 50m 19s   \\ \cline{3-8}
 \multirow{1}{*}{\makecell{ uVCIL-KDE-RBF}}             &      &      78.43K & 0d 00h 31m 36s   & 33.83K & 0d 00h 29m 21s & 103.81K & 1d 00h 43m 29s   \\ \hline
 \multirow{1}{*}{\makecell{ iCaRL \cite{iCaRL}}}                    & \multirow{2}{*}{\makecell{21.3M \\  (RestNet \cite{he2016deep})}}     &      \multirow{2}{*}{\bf{21.33M}} & 0d 17h 18m 12s  & \multirow{2}{*}{\bf{21.31M}} & 0d 20h 59m 14s  & \multirow{2}{*}{\bf{21.37M}} & 1d 23h 59m 23s (30 tasks) \\ 
\multirow{1}{*}{\makecell{ iCaRL+CL \cite{vCLIMB}}}              &      &         & 1d 23h 59m 25s & & 1d 23h 56m 37s & & 1d 23h 57m 58s (30 tasks) \\ \hline
\end{tabular}
}
\end{center}
\caption{The number of parameters and training time for UCF101, HMDB51, and SSv2.}
\label{tab:compational_cost}
\end{table}

\end{document}